\documentclass{article} 
\usepackage{iclr2027_conference,times}
\usepackage[T1]{fontenc}

\usepackage{amsmath,amsfonts,bm}

\def\eqref#1{equation~\ref{#1}}

\def\1{\bm{1}}

\DeclareMathAlphabet{\mathsfit}{\encodingdefault}{\sfdefault}{m}{sl}
\SetMathAlphabet{\mathsfit}{bold}{\encodingdefault}{\sfdefault}{bx}{n}

\usepackage{amsthm}
\newtheorem{lemma}{Lemma}

\newcommand{\gsComplexityBaseMax}{52}
\newcommand{\gsComplexityMax}{68}
\newcommand{\gsComplexityMin}{0.7}

\newcommand{\gsComplexityStructMax}{68}
\newcommand{\gsEffectMax}{1.00}
\newcommand{\gsEffectMedian}{0.63}
\newcommand{\gsEffectMin}{0.02}

\newcommand{\gsNcodec}{173}

\newcommand{\gsNfamilies}{29}

\newcommand{\gsNrule}{86}
\newcommand{\gsNseeds}{3}
\newcommand{\gsNskeletons}{3}
\newcommand{\gsNstruct}{77}
\newcommand{\gsNsynth}{26}
\newcommand{\gsNvariants}{336}
\newcommand{\gsVendiBehav}{5.1}
\newcommand{\gsVendiBehavSeeds}{2.2}
\newcommand{\gsVendiMech}{9.1}

\newcommand{\gsVendiMechSeeds}{2.5}

\usepackage{hyperref}
\usepackage{url}
\usepackage[table]{xcolor}
\usepackage{booktabs}
\usepackage{subcaption}
\usepackage{graphicx}
\usepackage[super]{nth}
\usepackage{multirow}
\usepackage{array}
\usepackage{placeins}  
\newcolumntype{L}[1]{>{\raggedright\arraybackslash}p{#1}}
\usepackage{xspace}
\usepackage{wrapfig}
\usepackage{tcolorbox}
\usepackage{dashrule} 
\newtcolorbox{taskexample}{colback=gray!5!white,colframe=gray!75!black,sharp corners,boxrule=0.4pt}
\usepackage{algorithm}
\usepackage{algorithmic}
\usepackage{amsmath,amsfonts,amssymb}
\usepackage{siunitx}
\usepackage{pifont}
\usepackage{xcolor}
\usepackage{soul}
\usepackage{enumitem}

\definecolor{posbg}{RGB}{210,245,210}
\definecolor{neubg}{RGB}{220,220,220}
\definecolor{negbg}{RGB}{255,200,200}

\definecolor{winbg}{RGB}{223,245,223}

\definecolor{postext}{HTML}{006400}
\definecolor{negtext}{HTML}{FF6B6B}
\definecolor{meanshift}{HTML}{003366}

\definecolor{correctgreen}{HTML}{e6ffed}
\definecolor{incorrectred}{HTML}{ffe6e6}
\definecolor{improvedblue}{HTML}{e6f2ff}

\newcommand{\gamesmith}{\textsc{GameSmith}\xspace}
\newcommand{\ourrl}{\textsc{Crab}\xspace} 
\definecolor{resultred}{HTML}{C00000}
\newcommand{\result}[1]{\textcolor{resultred}{[\if\relax\detokenize{#1}\relax TBD\else#1\fi]}}

\usepackage{tikz}
\usetikzlibrary{arrows.meta,positioning,fit,backgrounds,shapes.geometric}

\usepackage{fvextra}
\tcbuselibrary{skins,breakable,listings}

\lstdefinestyle{promptstyle}{basicstyle=\fontfamily{lmtt}\fontsize{7pt}{8.2pt}\selectfont,breaklines=true,breakautoindent=true,breakindent=1.5em,columns=fullflexible,keepspaces=true,showstringspaces=false,xleftmargin=0pt}
\newtcblisting{promptbox}[1]{breakable,enhanced,listing only,colback=gray!4!white,colframe=gray!55!black,boxrule=0.5pt,arc=1.5pt,left=3pt,right=3pt,top=3pt,bottom=3pt,title={#1},fonttitle=\small\bfseries,coltitle=white,colbacktitle=gray!55!black,listing style=promptstyle}
\title{GameSmith: Scaling Zero-Sum Self-Play with Synthesised Games}

\author{%
Weixian Waylon Li$^{1,*}$,\ \ Jiaxin Zhang$^{2,\dagger}$,\ \ Han Zhou$^{3}$,\ \ Chengzu Li$^{3}$,\ \ Ke Wang$^{1}$, \\
\bf Fengxiang He$^{1,\dagger}$,\ \ Yiwen Guo$^{4,\dagger}$ \\[2pt]
\normalfont\small
$^{1}$University of Edinburgh \ \ $^{2}$LIGHTSPEED \ \ $^{3}$University of Cambridge \ \ $^{4}$Independent Researcher \\
\normalfont\small\fontfamily{lmtt}\selectfont
waylon.li@ed.ac.uk \ \ jiaxijzhang@global.tencent.com \ \ fhe@ed.ac.uk \\ 
\normalfont\small\fontfamily{lmtt}\selectfont guoyiwen89@gmail.com
}

\iclrfinalcopy 
\begin{document}

\maketitle

\ificlrfinal
\renewcommand{\thefootnote}{\fnsymbol{footnote}}
\footnotetext[1]{This work was done during an internship of Weixian Waylon Li at LIGHTSPEED (UK).}
\footnotetext[2]{Corresponding authors.}
\renewcommand{\thefootnote}{\arabic{footnote}}
\fi

\begin{abstract}
Post-training large language models (LLMs) on two-player zero-sum games provides a verifiable reward, and the gains have been shown to transfer to general reasoning. Existing pipelines, however, rely on a handful of hand-written games and on advantage estimators tied to them, so both the supply of games and the advantage estimate on a large corpus limit scaling. We present \gamesmith, which scales both. An LLM synthesises games inside templates whose reward, observation, and hidden-information code is fixed. Each game is verified by seed-paired replay and by an independent re-implementation of its specification, then expanded into encoding, rule, and size variants, giving \gsNsynth{} new games in \gsNfamilies{} families and 336 verified variants from the three SPIRAL templates. The model is distilled from the self-play of a larger teacher (Qwen3-32B) on these games and then trained by self-play reinforcement learning with \ourrl, an advantage estimator that credits each decision against a shrinkage estimate of the expected return of its position and is defined on every variant. On Qwen3-4B-Base, distillation raises the 13-benchmark mean pass@1 from $45.5$ to $62.5$, above the three SPIRAL games at the same budget, and self-play with \ourrl{} reaches $64.9$. The ordering holds across models, and transfer grows with the number of game families. 
\end{abstract}

\section{Introduction}
\label{sec:intro}

Two-player zero-sum games are an attractive source of post-training data for large language models (LLMs), for three reasons: 
(1) the outcome of every game (win, loss, or draw), is computed by the game engine and serves as the reward, so the reward is verifiable;
(2) trajectories are generated by game play, so their supply is unbounded; and 
(3) because the same model plays both sides, the opponent improves together with it, which naturally leads to a curriculum process \citep{silver2017alphazero,baker2020emergent}. 
Post-training on games has been shown to transfer to mathematical and general reasoning \citep{cheng2024spag,liu2025spiral,yuan2025marshal}. SPIRAL \citep{liu2025spiral} and MARSHAL \citep{yuan2025marshal} demonstrate this with multi-turn self-play on three hand-written text games, and both identify the supply of new games as their limiting factor.
More broadly, across reinforcement learning (RL) on verifiable tasks, adding environments improves generalisation more reliably than adding compute on a fixed set \citep{stojanovski2025reasoninggym,rlve2025}.

Scaling game-based post-training, therefore, has two-fold requirements: 
(1) supply of games without relying on human labour. Although LLMs are able to write game code \citep{verma2025ggbench,faldor2024omniepic,envscaler2026}, existing pipelines check only that the generated code executes, not that it implements the intended game. 
Executability alone guarantees neither a correct implementation nor a trustworthy reward; and
(2) post-training that scales as the number of games grows, since under fixed budgets, the visits to a single game become sparse. 
Existing self-play methods estimate the advantage baseline separately for each game from its past returns \citep{liu2025spiral}, which rely on repeated playing on individual games. 

In this paper, we present \gamesmith, a framework that retains the setting of SPIRAL, namely multi-turn self-play on two-player zero-sum text games, and scales it on both fronts. To the best of our knowledge, \gamesmith{} is the first method in the literature that scales LLM post-training by self-play on synthesised two-player zero-sum games.

\begin{figure}[t]
    \centering
    \includegraphics[width=1.0\linewidth]{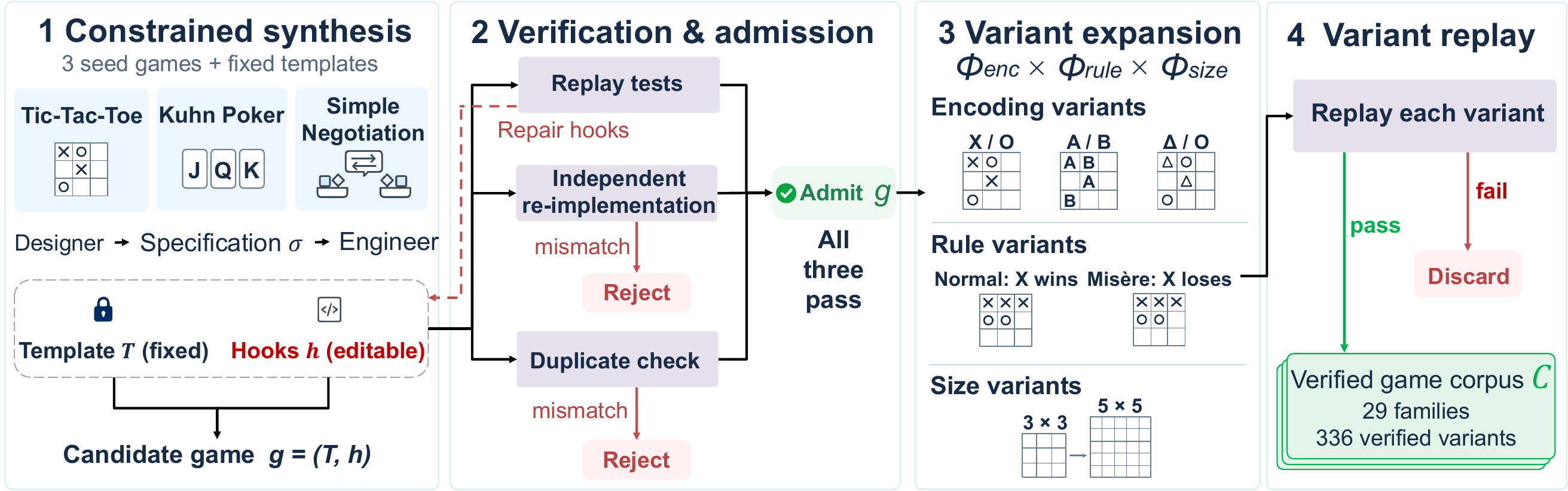}
    \caption{The \gamesmith{} game supply: an LLM fills the hooks of a fixed template, the game is admitted after replay tests, an independent re-implementation, and a duplicate check, and its encoding, rule, and size variants are replayed before entering the corpus.}
    \label{fig:overview}
    \vspace{-0.3cm}
\end{figure}

To scale the supply of games (Figure~\ref{fig:overview}), an LLM synthesises games inside \emph{templates}: the generated code fills in game-specific functions, while the code that assigns the reward, builds observations, and routes hidden information is fixed, so that these properties cannot be corrupted. 
The generated game logic is verified before use and then expanded into variants of its encoding, rules, and size. 
Building on the three templates derived from the SPIRAL games, this procedure creates \gsNsynth{} new games in \gsNfamilies{} families, with $163$ rule and size variants (\gsNrule{} and \gsNstruct{}) and \gsNcodec{} encoding variants. 

To scale post-training (Figure~\ref{fig:crab-rl}), \gamesmith{} trains its model in two stages, both using self-play trajectories from the synthesised games: (1) supervised fine-tuning on the self-play of a larger teacher model, Qwen3-32B, which plays the synthesised games against itself, followed by (2) reinforcement learning on the model's own self-play. For the second stage, we introduce \ourrl, a baseline for the policy-gradient estimate, i.e., the quantity subtracted from each return to reduce variance, that is defined on every corpus variant. \ourrl{} combines two estimates of the expected return of a game: the returns of the same game in the current batch, and a prior fitted across the corpus. The two are weighted as a Bayes estimator, and a state value computed by the template is then added for each position. Since it does not depend on the taken action, the policy-gradient estimate remains unbiased for the self-play objective. Further, under a standard uncorrelatedness assumption, its per-episode variance is lower than with either source alone.

We evaluate \gamesmith{} on $13$ reasoning benchmarks in mathematics, knowledge, logical reasoning, and spatial and state tracking. Distillation from the synthesised games raises Qwen3-4B-Base from $45.5$ to $62.5$ (mean pass@1 over four samples per question), outperforming both the same token budget of the three SPIRAL games ($60.6$) and the same teacher on budget-matched non-game reasoning data ($60.8$).
Self-play RL with \ourrl{} then further improves the average to $64.9$, outperforming the RAE algorithm proposed in SPIRAL.

\section{Related Work}
\label{sec:related}

\paragraph{Post-training on games.}
Self-play on games is the classic route to strong play \citep{silver2016alphago,silver2017alphazero}, and games have also been used to measure the strategic reasoning of LLMs \citep{duan2024gtbench,costarelli2024gamebench,bianchi2024negotiationarena}. As a source of post-training data, SPAG trains a policy through adversarial word games \citep{cheng2024spag}, SPIRAL runs multi-turn zero-sum self-play on hand-written text games \citep{liu2025spiral}, and MARSHAL extends this setting to several games \citep{yuan2025marshal}. All of these train on between one and six authored games and identify the supply of games as the bottleneck. A parallel line of work has the model write its own training problems and rewards it through code execution or agreement among its samples \citep{zhao2025absolutezero,huang2025rzero,liu2025spice,zhou2025selfchallenging,chen2025selfquestioning}, with self-play versions that pit a proposer against a solver or a critic \citep{kuba2025languageselfplay,chen2025spc}, and self-training retains the model's own samples that pass a verifier or a consistency check \citep{zelikman2022star,singh2024beyondhumandata,yuan2024selfreward,shafayat2025srt,li-etal-2023-bert}. In game post-training, the reward is the outcome of the game, which the proposer cannot manipulate.

\paragraph{Environment generation and variation.}
LLMs have been used to generate games and tasks as benchmarks \citep{verma2025ggbench}, as open-ended task pools \citep{faldor2024omniepic}, and as procedurally defined RL environments \citep{matthews2024kinetix,stojanovski2025reasoninggym}. Recent work further reports that the number of verifiable environments, rather than compute on a fixed set, is what drives generalisation \citep{rlve2025,envscaler2026,awm2026,dong2026agentworldscalingrealworldenvironme,li2026timelabelcontinuousphase}. 
The closest to our setting is GG-Bench \citep{verma2025ggbench}, where an LLM writes two-player games and $126$ of $1{,}000$ survive filters for execution, run time, and beatability by a trained agent, while the agreement between an implementation and its description is inspected by hand on ten of them. Generated environments have also been used for training: \citet{zhou2026verifiercurriculum} post-train a code generator by self-distillation on projects that pass an execution gate, \citet{shi2026evoenv} train a single policy by reinforcement learning on reasoning environments that it synthesises itself, admitted by execution, difficulty, and novelty checks, and \citet{jia2026gamegenverifier} verify generated games by reaching keypoints of the state from injected states. 
None of these involves two-player zero-sum games or self-play, and none checks an implementation against a second, independent reading of its specification. 
Unsupervised environment design and domain randomisation vary a single level space to improve robustness \citep{dennis2020paired,wang2019poet,tobin2017domain}, and procedural benchmarks measure generalisation across levels \citep{cobbe2019quantifyin}. 

\paragraph{Advantage baselines.}
Policy-gradient post-training of LLMs \citep{schulman2017ppo,ouyang2022instructgpt,deepseek2025r1,ICLR2026_32ddf8ee} typically uses group or leave-one-out baselines over samples of the same prompt \citep{shao2024deepseekmath,ahmadian2024backtobasics,liu2025drgrpo,yu2025dapo} or learned value estimates \citep{kazemnejad2025vineppo}. In game self-play, the analogue of a prompt is a pair of a variant and a seat, and SPIRAL maintains one running mean per such pair \citep{liu2025spiral}. 

\section{Preliminaries}
\label{sec:prelim}


\paragraph{Games.}
This paper studies two-player, turn-based, finite-horizon games with a zero-sum outcome: either one player wins and the other loses, or the game is drawn. 
The two positions a player can take are the \emph{seats} $o\in\{0,1\}$, where seat $0$ moves first. 
The \emph{trajectory} $\tau$ of a seat in an \emph{episode}, a single play of a game from start to end, is the sequence of observations and actions that the seat received and took, and its \emph{return} $R(\tau)\in\{+1,0,-1\}$ is the outcome from that seat's point of view, as computed by the game engine. 
A \emph{policy} maps an observation to a distribution over actions. 
We denote by $\pi_\theta$ the model being trained and by $\pi_{\mathrm{T}}$ the teacher from whose self-play it is distilled. 

\paragraph{Variants.}
A \emph{descriptor} $\phi=(\phi^{\mathrm{enc}},\phi^{\mathrm{rule}},\phi^{\mathrm{size}})$ specifies three components. 
The \emph{encoding} $\phi^{\mathrm{enc}}$ determines how a state is written (e.g., the marks, coordinate scheme, and orientation of a Tic-Tac-Toe board), and it changes the text but not the game. 
The \emph{rules} $\phi^{\mathrm{rule}}$ consist of a legality, transition, or outcome clause, such as forbidding the centre on the first move, and they change the game. 
The \emph{size} $\phi^{\mathrm{size}}$ specifies the size of the state and action spaces, such as a $5\times5$ board, and it changes the game while keeping its rules. 
A \emph{variant} $E_\phi$ is the game fixed to a descriptor, the \emph{family} of a game is the set of its admitted variants, and the \emph{corpus} is the union of all families. 
We write $\psi(\phi)=(\phi^{\mathrm{rule}},\phi^{\mathrm{size}})$ for the part of a descriptor that changes the game.
Encodings leave the game unchanged, but the model reads text and must learn this invariance from the renderings it observes, which is why they form a separate axis (Appendix~\ref{app:symmetry} and~\ref{app:encoding-efficiency}).

\paragraph{Policy gradient with a baseline.}
In self-play the same policy occupies both seats, and the update treats each seat in turn as the learner, with the opponent's actions held fixed as part of the environment. The objective of the learning seat is its expected return $J(\theta)=\mathbb{E}[R(\tau)]$ over variants and over trajectories $\tau$ of that seat, drawn from $\pi_\theta$ against a copy of the current policy through which no gradient is taken, and the update sums the gradients of the two seats. The objective is defined per seat because the return averaged over both seats of a zero-sum game is identically zero. Given a batch of $N$ seat trajectories $\tau_i$, the gradient is estimated by
\begin{equation}
\label{eq:pg}
\hat{g}\;=\;\frac{1}{N}\sum_{i=1}^{N}\big(R(\tau_i)-b_i\big)\,\nabla_\theta\log\pi_\theta(\tau_i),
\end{equation}
where the \emph{baseline} $b_i$ is any quantity that does not depend on the actions in $\tau_i$. Such a baseline leaves the estimate unbiased and affects only its variance \citep{ahmadian2024backtobasics,liu2025drgrpo}, and the variance is minimised when $b_i$ is closest in mean squared error to the expected return of the game and seat of episode $i$ (Appendix~\ref{app:proof}). In game self-play, $b_i$ must depend on the seat, since most games favour the first mover, and it must also be defined for every variant of a large corpus, most of which a run visits only a few times. RAE, the estimator of \citet{liu2025spiral}, maintains a running mean return per variant and seat and uses the corresponding cell as $b_i$.

\section{Method}
\label{sec:method}

\begin{wrapfigure}{r}{0.32\textwidth}
    \vspace{-0.3cm}
    \centering
    \includegraphics[width=\linewidth]{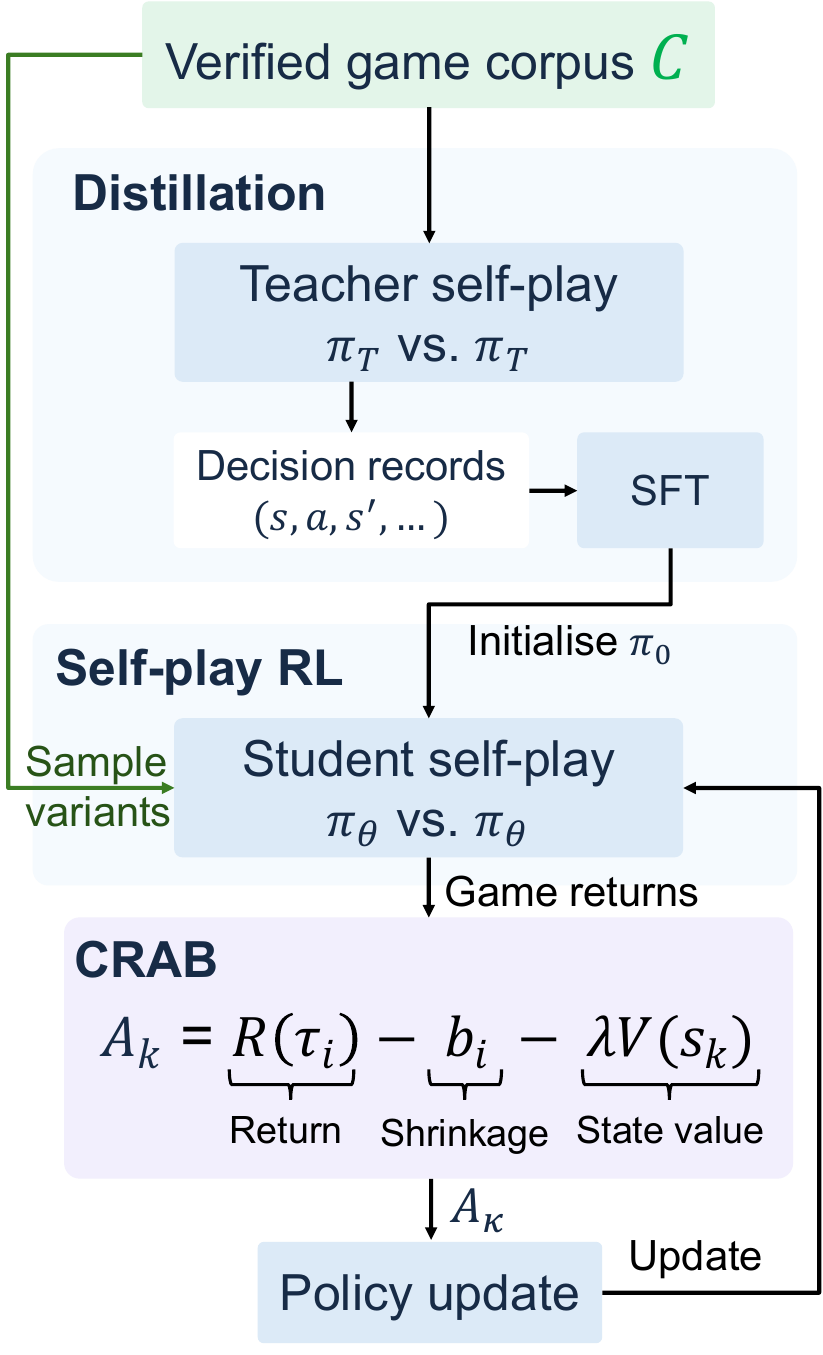}
    \caption{Post-training on the verified corpus with \ourrl{}.}
    \label{fig:crab-rl}
    \vspace{-0.5cm}
\end{wrapfigure}

\gamesmith{} scales game-based post-training on two fronts (Figures~\ref{fig:overview} and~\ref{fig:crab-rl}). On the data side, a synthesis pipeline lets an LLM write new zero-sum games and admits only those that pass verification (\S\ref{sec:method:synthesis}). On the post-training side, the resulting corpus has hundreds of variants that a run visits only a few times, which breaks the per-game advantage baselines of existing self-play methods, so we propose \ourrl{}, a baseline that is defined on every variant of the corpus (\S\ref{sec:method:ourrl}).

\subsection{Scaling the supply of games: synthesis with verification}
\label{sec:method:synthesis}

\paragraph{Templates.}
\label{sec:method:templates}

A \emph{template} $T=(K,H)$ consists of a \emph{kernel} $K$, which is code that is fixed for every game built on the template, and a \emph{hook set} $H=\{h_1,\dots,h_m\}$, which is the set of named functions that a game must supply. 
The kernel encodes observations, assigns rewards, and builds each seat's observation from the state, 
whereas the hooks return only game-specific verdicts: whether an action is legal, the next state, and who has won. 
A game is thus a template together with an implementation of its hooks, $g=(T,h)$. We use \gsNskeletons{} templates, one per genre, each seeded by one of the three SPIRAL games \citep{liu2025spiral}: grid placement with perfect information, betting with a showdown under imperfect information, and resource negotiation under private valuations. Each template has between one and four hooks (Appendix~\ref{app:hooks}) and settles every game as zero-sum.

Because the kernel is fixed, every game on a template has a zero-sum reward structure consistent with the hooks’ verdicts.
Hidden state reaches each seat only through the channels declared by the kernel
Encoding variants differ only in their text, since the kernel itself applies the bijection $e$ (\S\ref{sec:prelim}).

\paragraph{Generation.}
\label{sec:method:generation}

Games are written by two LLM roles. A \emph{designer} writes a specification consisting of the rules in natural language, the semantic parameters (e.g., board size, deck, code length), and the deductive skill that the game should exercise. 
An \emph{engineer} then implements the specification by filling in the hooks of the corresponding template, and it cannot modify the kernel. When a test fails, its output is returned to the engineer for revision, and a game that exhausts its bounded number of repair attempts is discarded (Appendix~\ref{app:prompts} gives the models, decoding, and prompts).

\paragraph{Verification.}
\label{sec:method:verification}

Every test is built on a single primitive, \emph{seed-paired replay}: two environments are run from the same random seed under the same driver, i.e., a policy that supplies actions to both, and a relation between the two trajectories is checked on every seed of a fixed set (Appendix~\ref{app:verify-tests} states each test formally as a predicate, and Appendix~\ref{app:verify:audit} reports which injected defects each test catches). 
Two tests have a known answer because of the kernel: an encoding variant must have identical outcomes once actions are mapped through $e$, and a rule variant must change the outcome on at least one seed. 
Three further tests check termination and zero-sum rewards, information hiding under perturbation of the hidden state, and playability under a reference bot (Table~\ref{tab:certs}).

Replay tests have not yet checked that the implementation is the game that the designer described, so a second model from a different model family implements the same specification without access to the first implementation, and the two implementations are replayed from shared seeds under the same driver. Any disagreement on an outcome leads to rejection of the game (Appendix~\ref{app:verify:audit} also compares ten admitted games with their specifications by hand). Finally, a game whose play statistics under the random driver are close to those of an already admitted game is replayed against it from shared seeds, and it is rejected as a duplicate if the outcomes match on every seed. A game is admitted when it passes every replay test, agrees with its re-implementation, and is not a duplicate. An LLM playtest then scores each admitted game on playability from the rules text alone, as a diagnostic that plays no role in admission.

\paragraph{Variant expansion.}
\label{sec:method:expansion}

Each admitted game is expanded into its family. The descriptor set is the product of three per-component sets,
\begin{equation}
\label{eq:phi}
\Phi(g)=\Phi_{\mathrm{enc}}(g)\times\Phi_{\mathrm{rule}}(g)\times\Phi_{\mathrm{size}}(g),
\end{equation}
where encodings are enumerated from the template's pools, rule variants are typed perturbations of the semantic parameters, and size variants change the state and action spaces. Every variant must itself pass the replay tests. Moreover, every rule and size variant is annotated with an \emph{effect size}, defined as the fraction of seed-paired episodes whose outcome differs between the variant and its parent. The variants used at each training stage are listed in Appendix~\ref{app:curation}.

\subsection{Scaling post-training: distillation and self-play with \ourrl{}}
\label{sec:method:ourrl}

The model is distilled from a teacher's self-play on the corpus and then trained by self-play RL, whose advantage baseline must be defined on every variant of a corpus that a run visits only a few times. We describe the distillation stage, then this baseline, and then the self-play stage.

\paragraph{Distillation.}
\label{sec:method:training}
\label{sec:method:distill}
The teacher $\pi_{\mathrm{T}}$ plays variants of every family against itself. Each decision is recorded as a prompt (rules text, current observation, and action format) and a response (a reasoning block followed by the chosen action). The student $\pi_\theta$ is then fine-tuned on these records.

\paragraph{Shrinkage baseline.}

\ourrl{}, a corpus-regularised advantage baseline, serves as the $b_i$ in \eqref{eq:pg}: it estimates the expected return of the game and seat of episode $i$ as a Bayes estimator that combines the information in the current batch with that of the rest of the corpus.

This quantity is keyed on what a variant is rather than on its identity. The \emph{group} of an episode is $g=(\psi(\phi),o)$, i.e., the components of the descriptor that change the game, together with the seat. The encoding is excluded because it is redrawn every episode and does not change the game (\S\ref{sec:prelim}). Consequently, the returns of one group are the returns of one game from one seat, averaged over its encodings. Let $\mu_g$ denote the expected return of group $g$ under the current policy. At each optimiser step, two sources of information about $\mu_g$ are available. The first is the batch itself. Because each sampled variant is played $G$ times (\S\ref{sec:method:training}), the batch contains $K_g\ge G$ episodes of group $g$, and the leave-one-out mean
\begin{equation}
\label{eq:loo}
\bar{y}_{g,-i}\;=\;\frac{1}{K_g-1}\sum_{j\in g,\,j\neq i} R(\tau_j)
\end{equation}
is an unbiased estimate of $\mu_g$ under the current policy, with variance $\sigma^2/(K_g-1)$, where $\sigma^2$ is the variance of a single return. The second is a \emph{prior} $m_g$, which is a linear model on the descriptor with one weight per value of each rule and size field, one weight per seat, and an intercept, fitted by online least squares on the returns of all earlier batches. The prior is defined for every descriptor in the corpus, including combinations of field values that no episode has visited, but it lags behind the policy and cannot express interactions between fields.

\ourrl{} combines the two sources as
\begin{equation}
\label{eq:ourrl}
b_i\;=\;\frac{K_g'\,\bar{y}_{g,-i}\;+\;\kappa\,m_g}{K_g'+\kappa},\qquad K_g'=K_g-1,
\end{equation}
where $\kappa>0$ is a pseudo-count that controls the weight of the prior. If the group means are modelled as $\mu_g\sim\mathcal{N}(m_g,\tau^2)$ and the returns of a group are modelled as independent with variance $\sigma^2$ about $\mu_g$, then \eqref{eq:ourrl} with $\kappa=\sigma^2/\tau^2$ is the posterior mean of $\mu_g$ given the other returns of the group. Without the normality assumption, it remains the linear estimator of $\mu_g$ with the smallest mean squared error given the same two moments. Intuitively, a group with many episodes in the batch ($K_g'\gg\kappa$) relies on its own mean, whereas a group with few episodes falls back on the prior. As $\kappa\to0$, $b_i$ reduces to the leave-one-out mean of RLOO \citep{ahmadian2024backtobasics} restricted to episodes of the same game and seat. As $\kappa\to\infty$, $b_i$ reduces to the prior alone. RAE corresponds to neither: its table has one cell per variant and seat, so it pools no information across variants.

The main runs fix $\kappa=4$, so that with $G=5$ the batch mean and the prior receive equal weight. The model also prescribes $\kappa$: $\sigma^2$ and $\tau^2=\mathbb{E}[(\mu_g-m_g)^2]$ are estimated from the returns of each batch, and $\hat{\kappa}=\hat{\sigma}^2/\hat{\tau}^2$ is recorded for every run, together with $\hat{\tau}^2_{\mathrm{enc}}$, the same variance across encodings within a group, which is zero for an equivariant policy (Appendix~\ref{app:error} gives the estimators, a variant of \ourrl{} that uses $\hat{\kappa}$, and the error analysis).

The baseline $b_i$ depends only on the other episodes of the batch and on a prior fitted before the batch, and not on the actions of $\tau_i$. Therefore, \eqref{eq:pg} remains unbiased for any $\kappa$, whether fixed or estimated with lag. Its variance is governed by how well $b_i$ estimates $\mu_g$.
\begin{lemma}
\label{prop:variance}
Let $\hat{g}_i=(R(\tau_i)-b_i)\nabla_\theta\log\pi_\theta(\tau_i)$ be the term of \eqref{eq:pg} contributed by episode $i$, and assume that the squared norm of the score $\nabla_\theta\log\pi_\theta(\tau_i)$ is uncorrelated with the squared return residual $(R(\tau_i)-b_i)^2$, which is the assumption under which the variance-minimising constant baseline is the expected return. Then $\mathrm{Var}[\hat{g}_i]$ is an increasing affine function of $\mathbb{E}[(b_i-\mu_g)^2]$. Under the model above, this error equals $\sigma^2/(K_g'+\kappa^\star)$ at $\kappa^\star=\sigma^2/\tau^2$, which is smaller than both the error $\sigma^2/K_g'$ of the leave-one-out mean alone and the error $\tau^2$ of the prior alone.
\end{lemma}
The proof is given in Appendix~\ref{app:proof}. Under the stated assumption, the estimator therefore reduces the variance of each gradient term relative to both of its limiting cases when $\kappa$ is close to $\sigma^2/\tau^2$, which is precisely the value that $\hat{\kappa}$ targets.

\paragraph{Self-play.}
\label{sec:method:rl}
In the second stage, the distilled model $\pi_\theta$ plays against a copy of itself. Each optimiser step samples variants and plays each one $G$ times with fresh seeds. The encoding is drawn afresh for every episode and held fixed within the episode, so that a batch contains $G$ episodes of every sampled game under different encodings. A drawn episode returns $0$ to both seats and is treated as a sample of \eqref{eq:pg} like any other, and draws remain rare on the corpus (Appendix~\ref{app:draws}). The policy is updated with \eqref{eq:pg}, where the $k$-th decision of $\tau_i$, taken in state $s_k$, receives the advantage
\begin{equation}
\label{eq:ourrl-turn}
A_k \;=\; R(\tau_i)\;-\;b_i\;-\;\lambda\,V(s_k),
\end{equation}
with $b_i$ given by \eqref{eq:ourrl}, $V(s_k)\in[-1,1]$ a value of position $s_k$ for the seat to move, which is a fixed function of the information visible to that seat that the template computes from its own hooks (Appendix~\ref{app:hooks} defines it for each template), and $\lambda>0$ a fixed weight. Both terms are absolute levels: $b_i$ is the expected return of the game from this seat and $V(s_k)$ the value of the position under a fixed reference behaviour, which varies within the episode. Neither term depends on the action taken at decision $k$, so $b_i+\lambda V(s_k)$ is an action-independent baseline for that decision, and \eqref{eq:ourrl-turn} keeps the gradient unbiased for any $V$.

\section{Experiments}
\label{sec:experiments}

We conducted extensive experiments addressing four research questions. \textbf{RQ1:} At a fixed teacher and token budget, does distillation from synthesised games transfer to general reasoning better than distillation from the three SPIRAL games, from non-game reasoning data, or from game data with the reasoning removed? \textbf{RQ2:} Does self-play on the synthesised corpus improve on the distilled model, does the full pipeline exceed the SPIRAL pipeline, and does \ourrl{} improve on RAE? \textbf{RQ3:} How does transfer scale with the number of game families and with the number of trajectories? \textbf{RQ4:} Which properties of the corpus and which capabilities of the model account for the transfer?

\subsection{Implementation details}
\label{sec:setup}

\textbf{Corpus.} The corpus is seeded by the three games of SPIRAL \citep{liu2025spiral}, implemented on TextArena \citep{guertler2025textarena}, with one game per template: Tic-Tac-Toe, Kuhn Poker, and Simple Negotiation. The synthesiser admits \gsNsynth{} new games, resulting in \gsNfamilies{} families with $163$ rule and size variants that change the game (\gsNrule{} and \gsNstruct{}) and \gsNcodec{} encoding variants that do not (Appendix~\ref{app:games}).

\textbf{Models.} Our main model is Qwen3-4B-Base \citep{yang2025qwen3}, and the main comparisons are repeated on Qwen3-8B-Base and on two instruct models from other families, Mistral-Nemo-Instruct-2407 and GLM-4-9B-0414 \citep{glm2024chatglm}. The teacher for every distillation setting is Qwen3-32B.

\textbf{Pipeline.} 
Distillation data are built as in \S\ref{sec:method:training}, with the action in \texttt{\textbackslash boxed}, and self-play with \ourrl{} then starts from the distilled checkpoint. 
The hyperparameters, hardware, software, and compute are described in Appendices~\ref{app:hparams} and~\ref{app:compute}.

\textbf{Implementations.} 
All distillation settings share the same teacher, decoding configuration, output format, and token budget, and differ only in the environments that the teacher plays. 
SPIRAL distills from the three SPIRAL games only and serves as the main baseline. 
For \gamesmith{}, we subsample our corpus to the same token budget as the baselines ($8{,}208$ trajectories). 
Non-game uses the same teacher on non-game reasoning problems (deductive logic, spatial and state tracking, and competition mathematics) at the same budget, decontaminated against every evaluation set (Appendix~\ref{app:teacher-data}). 
Format keeps the \gamesmith{} prompts and the boxed action but removes the reasoning. For self-play RL, the SPIRAL pipeline runs the same two stages on the three SPIRAL games with RAE, the estimator of \citet{liu2025spiral}, which amounts to SPIRAL with a distillation stage in front. 
The estimator control keeps the \gamesmith{} corpus and checkpoint, the group size $G=5$, and the token budget, and replaces \ourrl{} with RAE. 

\textbf{Evaluation.} We report pass@1 over $4$ samples per question with an $8{,}192$-token generation budget (Appendix~\ref{app:eval}) on $13$ benchmarks in four groups: mathematics (MATH-500 \citep{hendrycks2021math,lightman2023verify}, AIME \citep{sun-etal-2026-challenging}, AMC\footnote{\url{https://huggingface.co/datasets/AI-MO/aimo-validation-amc}}, Minerva Math \citep{lewkowycz2022minerva}, and OlympiadBench \citep{he2024olympiadbench}, the five benchmarks used by \citet{liu2025spiral}); knowledge (MMLU-Pro \citep{wang2024mmlupro} and GPQA-Diamond \citep{rein2024gpqa}); logical reasoning (BBH Logical Deduction with three and five objects and Web of Lies \citep{suzgun2023bbh}); and spatial and state tracking (BBH Tracking Shuffled Objects, Geometric Shapes, and Navigate). The headline number is the unweighted mean over the $13$ benchmarks.

\subsection{Empirical results}
\label{sec:results}

\paragraph{RQ1: Distillation from synthesised games.}
\label{sec:results:sft}

Table~\ref{tab:general-reasoning} compares the distillation settings on four models (per-benchmark results in Appendix~\ref{app:full-tables}). We make two observations. 
First, the synthesised games are the strongest source of reasoning data on three of the four models. 
On Qwen3-4B-Base, \gamesmith{} (matched) raises the 13-benchmark mean from $45.5$ to $62.5$, compared with $60.8$ for Non-game, $60.6$ for SPIRAL, and $28.8$ for Format (Appendix~\ref{app:format}). 
Qwen3-8B-Base and GLM-4-9B-0414 exhibit the same ordering. 
On Mistral-Nemo-Instruct-2407, where game play transfers weakly, non-game reasoning data at the same token budget transfers better, by $5.8$ points, and both game-based settings lose ground on GPQA-Diamond (Table~\ref{tab:full-mistralnemoinstruct2407}). 
Note that the Non-game control is built from eight annotated datasets, keeps only traces that match the gold answer, and is $43\%$ competition mathematics (Appendix~\ref{app:teacher-data}). 
In contrast, \gamesmith{} uses no annotation of any kind, yet on Qwen3-4B-Base it still matches that corpus on logical reasoning ($83.4$ against $84.1$) and outperforms it on mathematics ($45.7$ against $40.6$), a capability that the games do not contain at all. 
The other two models on which game play transfers behave in the same way. Beyond needing no annotation, game data carries an engine-computed reward, so the same corpus drives the self-play stage of Table~\ref{tab:pipelines}, where annotated data has no further use, and its supply grows on demand (RQ3). 
Second, compared with SPIRAL, the control that changes only which games the teacher plays, the synthesised games are worth more per token on every model, by between $1.9$ and $5.2$ points. 
Synthesised games are therefore the stronger source of game data on every model, and the stronger source of reasoning data on all but the weakest of them. 

\begin{table}[t]
\centering\small
\setlength{\tabcolsep}{3pt}
\caption{Distillation results across models (pass@1 \%, $8{,}192$-token budget): capability-group means and the 13-benchmark mean. All SFT settings share the same teacher (Qwen3-32B), format, and token budget. Bold indicates the best value in each row. A shaded cell indicates that \gamesmith{} exceeds the SPIRAL baseline on that row.}
\begin{tabular}{@{}ll ccccc@{}}
\toprule
& & & \multicolumn{3}{c}{Controls} & \gamesmith{} \\
\cmidrule(lr){4-6}\cmidrule(lr){7-7}
Model & Capability & Base & Format & Non-game & SPIRAL & matched \\
\midrule
Qwen3-4B-Base & Mathematics & 32.5 & 10.7 & 40.6 & 39.6 & \cellcolor{winbg}\textbf{45.7} \\
 & Knowledge & 42.9 & 34.5 & 47.4 & 48.3 & \cellcolor{winbg}\textbf{49.6} \\
 & Logical reasoning & 54.9 & 50.9 & 84.1 & \textbf{85.5} & 83.4 \\
 & Spatial \& tracking & 59.5 & 33.1 & \textbf{80.4} & 78.8 & 78.4 \\
\cmidrule{2-7}
 & Average (13 benchmarks) & 45.5 & 28.8 & 60.8 & 60.6 & \cellcolor{winbg}\textbf{62.5} \\
\midrule
Qwen3-8B-Base & Mathematics & 37.4 & 7.2 & 44.4 & 50.5 & \cellcolor{winbg}\textbf{51.6} \\
 & Knowledge & 48.5 & 4.3 & 55.4 & 56.9 & \cellcolor{winbg}\textbf{58.4} \\
 & Logical reasoning & 78.1 & 25.0 & 94.7 & 90.8 & \cellcolor{winbg}\textbf{95.4} \\
 & Spatial \& tracking & 59.0 & 17.6 & \textbf{85.6} & 83.4 & \cellcolor{winbg}84.2 \\
\cmidrule{2-7}
 & Average (13 benchmarks) & 53.5 & 13.3 & 67.2 & 68.4 & \cellcolor{winbg}\textbf{70.3} \\
\midrule
Mistral-Nemo-Instruct-2407 & Mathematics & 16.1 & 2.1 & \textbf{19.6} & 12.3 & \cellcolor{winbg}17.2 \\
 & Knowledge & 34.1 & 2.3 & \textbf{41.2} & 28.4 & \cellcolor{winbg}33.4 \\
 & Logical reasoning & 70.0 & 14.5 & \textbf{93.9} & 76.5 & \cellcolor{winbg}80.3 \\
 & Spatial \& tracking & 66.9 & 16.9 & \textbf{78.4} & 69.0 & \cellcolor{winbg}76.2 \\
\cmidrule{2-7}
 & Average (13 benchmarks) & 43.1 & 8.4 & \textbf{53.7} & 42.7 & \cellcolor{winbg}47.9 \\
\midrule
GLM-4-9B-0414 & Mathematics & 32.2 & 2.7 & 34.5 & 38.5 & \cellcolor{winbg}\textbf{40.3} \\
 & Knowledge & 48.1 & 12.4 & 48.1 & 48.2 & \cellcolor{winbg}\textbf{50.1} \\
 & Logical reasoning & 86.3 & 33.5 & 94.5 & 92.8 & \cellcolor{winbg}\textbf{95.0} \\
 & Spatial \& tracking & 63.4 & 30.5 & 83.8 & 82.6 & \cellcolor{winbg}\textbf{86.4} \\
\cmidrule{2-7}
 & Average (13 benchmarks) & 54.3 & 17.7 & 61.8 & 62.7 & \cellcolor{winbg}\textbf{65.1} \\
\bottomrule
\end{tabular}
\label{tab:general-reasoning}
\end{table}

\paragraph{RQ2: Self-play on top of distillation.}
\label{sec:results:selfplay}

Table~\ref{tab:pipelines} reports three self-play pipelines: SPIRAL's three games with RAE from the SPIRAL distillation checkpoint, which is SPIRAL's original recipe, the synthesised corpus with RAE from the \gamesmith{} checkpoint, and \gamesmith{}, i.e., the same corpus and checkpoint with \ourrl{}. The first two differ in both the checkpoint and the games and therefore compare complete pipelines. The last two share both and differ in the estimator (RAE vs. \ourrl{}). 
\ourrl{} improves on the distilled model and exceeds RAE on every model. On Qwen3-4B-Base, self-play with \ourrl{} raises the mean from $62.5$ to $64.9$, whereas RAE on the same corpus and checkpoint reaches only $62.7$ and the SPIRAL pipeline reaches $62.5$ from $60.6$. The other three models give the same ordering, and on Mistral-Nemo-Instruct-2407 and GLM-4-9B-0414 RAE falls below its starting point ($47.9$ to $47.3$ and $65.1$ to $64.9$) while \ourrl{} stays above it. The gain from self-play is smallest on GLM-4-9B-0414 ($+0.3$), where distillation already saturates the model at $65.1$, so there is little left for self-play to add. RAE on the synthesised corpus is also unstable: after an early peak its scores fall well below the starting point for most of the budget, so its best checkpoint typically lies within the first $40$ steps (Appendix~\ref{app:rl-curves}). A comparison of the two estimators with one episode per variant is reported in Appendix~\ref{app:group-size}. Across models, the full \gamesmith{} pipeline exceeds the full SPIRAL pipeline by $1.6$ to $5.9$ points, and \ourrl{} exceeds RAE on the same corpus by $0.5$ to $2.2$ points (win rates on held-out game variants are reported in Appendix~\ref{app:game-strength}).

\begin{table}[t]
\centering\small
\setlength{\tabcolsep}{4pt}
\caption{Self-play post-training results across models (13-benchmark mean pass@1 \%).}
\begin{tabular}{@{}lcccccc@{}}
\toprule
& & \multicolumn{2}{c}{SPIRAL games} & \multicolumn{3}{c}{\gamesmith{} games} \\
\cmidrule(lr){3-4}\cmidrule(lr){5-7}
Model & Base & Distill. & +RL (RAE) & Distill. & +RL (RAE) & +RL (\ourrl{}) \\
\midrule
Qwen3-4B-Base & 45.5 & 60.6 & 62.5 & 62.5 & 62.7 & \textbf{64.9} \\
Qwen3-8B-Base & 53.5 & 68.4 & 68.8 & 70.3 & 72.0 & \textbf{73.1} \\
Mistral-Nemo-Instruct-2407 & 43.1 & 42.7 & 42.9 & 47.9 & 47.3 & \textbf{48.8} \\
GLM-4-9B-0414 & 54.3 & 62.7 & 63.8 & 65.1 & 64.9 & \textbf{65.4} \\
\bottomrule
\end{tabular}
\label{tab:pipelines}
\end{table}

\paragraph{RQ3: Scaling the number of families and the number of trajectories.}
\label{sec:results:scaling}

\begin{figure}[t]
\centering
\includegraphics[width=1.0\linewidth]{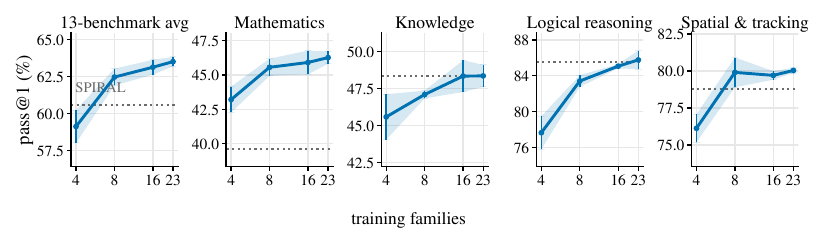}\\
\vspace{-0.2cm}
\includegraphics[width=1.0\linewidth]{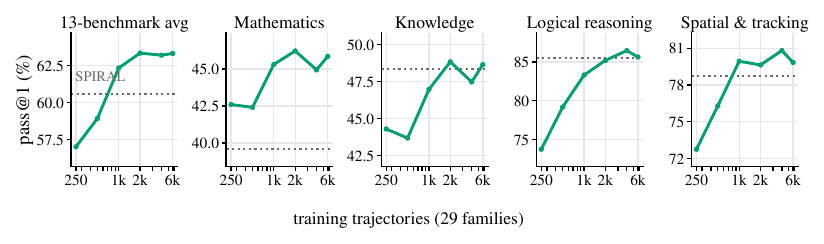}
\vspace{-0.5cm}
\caption{Distillation transfer by capability group on Qwen3-4B-Base (pass@1 \%). \textbf{Top}: scaling the number of families ($4\to23$) at a fixed number of trajectories, showing the mean and one standard deviation over three draws of families. \textbf{Bottom}: scaling the number of trajectories with all \gsNfamilies{} families. Dotted lines mark the SPIRAL performance.}
\label{fig:rq3-envscaling}
\vspace{-0.4cm}
\end{figure}

A larger corpus can mean more game families or more trajectories, and Figure~\ref{fig:rq3-envscaling} separates the two on Qwen3-4B-Base: transfer increases with both. 
With the number of trajectories fixed, the mean increases from $59.1$ at $4$ families to $63.5$ at $23$, with the largest step between $4$ and $8$ families, and every capability group improves, with logical reasoning improving the most. 
With all \gsNfamilies{} families, the mean rises steeply up to about $1{,}000$ trajectories ($57.0\to62.3$) and remains flat thereafter: $1{,}000$ trajectories already exceed the SPIRAL performance, and the matched budget of Table~\ref{tab:general-reasoning} lies on the flat part of the curve, so the benefit comes from covering more games and is reached with few trajectories.

\paragraph{RQ4: Corpus diversity and which capabilities transfer.}
\label{sec:analysis}

\label{sec:analysis:corpus}The corpus is more diverse than the three SPIRAL games and spans a wider range of game-tree complexity (Appendix~\ref{app:diversity}): the Vendi score by mechanic is \gsVendiMech{} against \gsVendiMechSeeds{}, a rule change flips between \gsEffectMin{} and \gsEffectMax{} of outcomes, and game-tree complexity spans $[\gsComplexityMin{},\gsComplexityMax{}]$ across the \gsNvariants{} variants, supplied by the size variants.

\label{sec:analysis:capabilities}When read by capability group, Tables~\ref{tab:general-reasoning} and~\ref{tab:pipelines} and Figure~\ref{fig:rq3-envscaling} agree on what game data teaches. Distillation improves most the two capabilities that the games exercise most: multi-step deduction over stated constraints (logical reasoning, $+28.5$ from Base) and tracking a state through a sequence of moves (spatial and state tracking, $+18.9$). Mathematics improves less ($+13.2$) and knowledge the least ($+6.7$). 
Self-play sharpens the same profile: the RL gain is concentrated on logical reasoning ($+6.5$ on Qwen3-4B-Base), where the deduction tasks most closely resemble the game decisions for which the policy is rewarded. 
The increment of the synthesised games over the SPIRAL games has a different profile: on Qwen3-4B-Base it is concentrated on mathematics ($+6.1$), with the two groups the games exercise most at or slightly below SPIRAL ($-2.1$ and $-0.4$), and on the other three models it is positive in every group (Table~\ref{tab:general-reasoning}). What the synthesised games add over the seed games is therefore not confined to the capabilities that game data teaches best. Two features of the corpus account for this: the seed games already exercise deduction and state tracking, where the SPIRAL setting reaches $85.5$ and $78.8$ on Qwen3-4B-Base, so more games of the same kind add little, whereas the betting and negotiation families score hands and inventories by arithmetic (sums modulo $M$, bitwise XOR, target sums, set and tier bonuses, Table~\ref{tab:games}), so their teacher traces carry numerical computation that the seed games do not require. 

\section{Conclusion}
\label{sec:conclusion}

We presented \gamesmith{}, a framework for scaling LLM post-training with synthesised two-player zero-sum games. It combines constrained game synthesis and verification with distillation of teacher self-play and self-play reinforcement learning, and its advantage estimator, \ourrl{}, shares information across game variants where observations of individual variants are limited. Across four models and $13$ reasoning benchmarks, distillation on the synthesised games outperforms distillation on the SPIRAL games under matched token budgets, self-play with \ourrl{} adds up to $2.8$ points on top of distillation and exceeds both RAE on the same corpus and the SPIRAL pipeline on every model, and transfer improves as more game families are included, with a significant improvement already at $1{,}000$ trajectories. These findings support verified game synthesis as a scalable source of distillation data and reinforcement learning environments whose rewards come directly from game outcomes rather than from annotation.

\section*{Reproducibility statement}
The templates and their hook contracts are provided in Appendix~\ref{app:hooks}, the verification tests with their drivers, seeds, and relations in Appendix~\ref{app:verify}, the admitted games and the variants used at each training stage in Appendix~\ref{app:games} and~\ref{app:curation}, the proof of Lemma~\ref{prop:variance} in Appendix~\ref{app:proof}, the hyperparameters, software versions, hardware, and evaluation protocol in Appendices~\ref{app:hparams} and~\ref{app:eval}, the licences of every asset in Appendix~\ref{app:licences}, the per-benchmark results in Appendix~\ref{app:full-tables}, and the verbatim prompts of the synthesis pipeline in Appendix~\ref{app:prompts}. The corpus, the synthesis and verification code, the training configurations, and the evaluation harness will be released upon acceptance.

\bibliography{iclr2027_conference}
\bibliographystyle{iclr2027_conference}

\newpage
\appendix
\section{Additional details of the method}
\label{app:method}
\subsection{Template hooks and state values}
\label{app:hooks}
Table~\ref{tab:hooks} lists, for each template of the main study, the functions that the engineer must or may implement and what each of them returns. Everything else is part of the kernel, including dealing and hidden-state routing, the betting or offer state machine, board rendering, action parsing, the encoding $e$, reward assignment, and episode termination. Hooks receive canonical indices (player ids $0$ and $1$, canonical coordinates, rank indices, and resource indices) rather than display symbols, and they must be deterministic and free of I/O. The rules text is the only hook that produces text for the player, and it must describe the game through a dictionary of the current display labels, so that a change of encoding changes the text consistently. The state value $V(s)$ of \S\ref{sec:method:rl} is computed by the kernel from these hooks and uses only information visible to the seat being valued, never the opponent's hidden state. For grid placement, it is the minimax outcome when the position is decided within two plies, and otherwise the mean outcome of $16$ random playouts whose seed is derived from the position, so that $V$ is a deterministic function of the state. For betting with a showdown, it combines the probability that the seat's own hand beats a random opponent hand drawn from the remaining deck, computed by exact enumeration and mapped to $[-1,1]$, with the settled chip lead normalised by the largest possible chip total, weighted by the fraction of rounds remaining. For resource negotiation, it is the gain of the seat's own inventory score over its initial value under the seat's private valuation, normalised by the scale of a single trade.

\begin{table}[ht]
\centering\small
\setlength{\tabcolsep}{4pt}
\caption{Engineer hooks per template. Each template also exposes a state value $V(s)$ (\S\ref{sec:method:ourrl}) computed by the kernel from these hooks.}
\begin{tabular}{@{}l L{0.33\linewidth} L{0.41\linewidth}@{}}
\toprule
Template & Hook and signature & Contract \\
\midrule
Grid placement
 & \texttt{legal\_moves(board, pid, extras)} $\to$ list of cells & optional, default is every empty cell \\
\addlinespace
 & \texttt{apply\_move(board, pid, r, c, extras)} & optional, default writes \texttt{pid} into the cell \\
\addlinespace
 & \texttt{outcome(board, pid, r, c, extras)} & called after each placement, returns none, a win or loss for a player, or a draw \\
\addlinespace
 & \texttt{rules\_text(cfg, player\_id, symbols)} $\to$ str & rules in natural language using the two current symbols \\
\addlinespace
 & \texttt{init\_extras()} $\to$ dict & optional per-episode state \\
\midrule
Betting with showdown
 & \texttt{hand\_value(hand, cfg)} $\to$ comparable & strength of a hand of rank indices, where the higher hand wins the showdown and equal values split the pot \\
\addlinespace
 & \texttt{rules\_text(cfg, player\_id, labels)} $\to$ str & rules using the current action words and rank labels \\
\midrule
Resource negotiation
 & \texttt{inventory\_score(}\allowbreak\texttt{quantities, values, cfg)} $\to$ int & score of an inventory under the player's private unit values, default $\sum_i q_i v_i$ \\
\addlinespace
 & \texttt{menu(cfg)} $\to$ list of swaps & optional, the proposable trades as (give, get, give quantity, get quantity) in a stable order, default every ordered pair at base-value parity \\
\addlinespace
 & \texttt{trade\_ok(give, get, gq, rq, quantities, cfg)} $\to$ bool & optional, default requires the proposer to hold the quantity given \\
\addlinespace
 & \texttt{rules\_text(cfg, player\_id, labels)} $\to$ str & rules using the current resource names, base values, and action words \\
\bottomrule
\end{tabular}
\label{tab:hooks}
\end{table}

\FloatBarrier

\subsection{Definition of the verification tests}
\label{app:verify}
\label{app:verify-tests}
Every replay test of \S\ref{sec:method:verification} is a predicate of the same form. Formally, let $\rho(E,s,a_{1:t})$ denote the trajectory of environment $E$ started from seed $s$ under the action sequence $a_{1:t}$. A test $c$ specifies which two environments to replay, a driver $D_c$, and a relation $R_c$ that the two trajectories must satisfy on every seed of a fixed set $\mathcal{S}$:
\begin{equation}
\label{eq:replay-test}
c(g)\;\equiv\;\forall\,s\in\mathcal{S},\ \forall\,a_{1:t}\sim D_c:\ R_c\big(\rho(E_1,s,a_{1:t}),\,\rho(E_2,s,a_{1:t})\big).
\end{equation}
Table~\ref{tab:certs} lists the replay tests in this form, together with the driver, the number of seeds, and the relation checked, followed by the agreement check against the independent re-implementation and the duplicate check.

\begin{table}[h]
\centering\footnotesize
\setlength{\tabcolsep}{4pt}
\caption{Replay tests, agreement check, and duplicate check as implemented. Each row is a predicate of the form of \eqref{eq:replay-test}.}
\begin{tabular}{@{}L{0.11\linewidth} r L{0.52\linewidth} L{0.20\linewidth}@{}}
\toprule
Test & Seeds & What is replayed, and what must hold on every seed & Property established \\
\midrule
Encoding & 40 & Random legal actions in $E_\phi$, mapped through $e$ into an encoding variant $E_{\phi'}$. The sets of available actions must correspond under $e$ at every step, and the final rewards must be identical. & hooks do not depend on the encoding \\
\midrule
Rule & 60 & The same random action sequence in $E_\phi$ and in its rule variant. For an outcome-reversing rule every decisive outcome must be swapped, otherwise some outcome must differ, and for a legality rule the first-turn set of legal actions must differ. & the rule change is effective \\
\midrule
Outcome & 60 / 20 / 1 & One variant under random play, a driver that submits illegal actions, and every rendered action label of the first turn. The two rewards must sum to zero and the episode must terminate, an illegal action must be penalised at once without stopping the engine, and every label must be accepted. & rewards are zero-sum and follow the hook verdict, termination, non-degeneracy \\
\midrule
Information hiding & 20 & The same action sequence in $E_\phi$ and in a copy with the opponent's hidden cards, code, or private unit values replaced. The observation stream of the first player must be identical up to the showdown or reveal. & hooks do not leak hidden state \\
\midrule
Playability & 40 & One variant with each seat driven by a reference bot maximising its own score. The bot must use the game's core action (for negotiation, at least $0.5$ executed trades per episode) and at least $25\%$ of episodes must be decisive. & the game can be played \\
\midrule
Agreement ($\mathrm{Agree}$) & 60 & The same labelled action sequence in $g$ and in $\tilde g$. The final rewards must be identical. & the two implementations agree on every replayed outcome \\
\midrule
Duplicate ($\mathrm{Dup}$) & 60 & The same action sequence in $g$ and in a fingerprint-close admitted game. The game is rejected if the outcomes match on every seed. & the game is new \\
\bottomrule
\end{tabular}
\label{tab:certs}
\end{table}

Both the duplicate check and the degeneracy flag rely on a fingerprint of random play. Each game is fingerprinted from $80$ episodes of random play by its mean episode length, mean branching factor, fraction of decisive outcomes, first-mover win share, and, for grid games, the fraction of the board filled. Two games on the same template whose fingerprints are close (episode length within $2$, decisive fraction within $0.1$, and fill fraction within $0.1$) are replayed against each other from $60$ shared seeds, and the later one is rejected if the outcomes match on every seed. The same fingerprint flags a game as degenerate, and thereby excludes it from training, if it produces only draws, if the first-mover win share lies outside $[0.1, 0.9]$, or if a grid game ends in fewer than four moves on average.

The independent re-implementation of each specification is written by a locally served Qwen3-32B, which belongs to a different model family from the GPT-5 generator, at temperature $0.6$ with up to $10{,}000$ completion tokens and at most three attempts. It is prompted with the engineer system prompt of the same template and receives the template source, the specification, and the configuration defaults of the first implementation, but never the first implementation itself. Before the agreement check, the re-implementation must itself pass the outcome test of Table~\ref{tab:certs}.

In the LLM playtest, a model plays one seat from the rules text and the current observation alone, and states with each move whether the game ends for it according to the rules. Its scores are the fraction of legal moves, the win rate against a random opponent, and the precision of those end-of-game claims against the engine. These scores are recorded as a diagnostic, a family whose legality score is below $0.7$ is flagged, and no score plays a role in admission. The prompt is given in Appendix~\ref{app:prompts}.

The replay tests are calibrated on the three SPIRAL games. Each game is first run through the tests unmodified, and every test must pass. Known defects, including a reversed outcome and a corrupted action label, are then injected into its hooks, and the tests are run again. 

One failure mode passed the first four tests and motivated the playability test. A negotiation family whose inventory score is strongly non-linear passes the outcome and information-hiding tests, and its random-play fingerprint is unremarkable, because random players do trade. Under a score-maximising reference bot, however, every swap on the default parity menu loses value for at least one side, so the bot never trades and every game ends in a draw. This failure mode is the reason why the engineer prompt for this template asks for a menu that is at parity under the game's own scoring rule.

\subsection{Effectiveness of the verification tests}
\label{app:verify:audit}

Table~\ref{tab:certs} lists the tests. In this subsection, we report what they did during synthesis and what they can be shown to catch. Specifically, we report three things: the candidates that each test eliminated during synthesis, the defects that each test catches when known defects are injected into the three seed games, and whether either result depends on the fixed seeds of Table~\ref{tab:certs}.

\paragraph{What each test eliminated.}
Table~\ref{tab:gate-ledger} aggregates the admission logs of every synthesis run. The designer proposed $80$ games, of which $76$ were implemented within the attempt budget and $75$ passed the replay tests, at $1.74$ engineer attempts per game (Appendix~\ref{app:games}). Four of the seven tests rejected no candidate in this batch. Since a test that never fires provides no evidence of its own efficacy, this motivates the injection experiment below. Across the $613$ certificate records stored with the corpus, none records a failure: every rejection that did occur is recorded separately, as a rejection record, a disagreement record, or a run report.

\begin{table}[h]
\centering\footnotesize
\setlength{\tabcolsep}{5pt}
\caption{Candidates eliminated by each test across all synthesis runs. Ran on is the number of families with a stored record for the test, and Rejected the number it turned away.}
\begin{tabular}{@{}l r r L{0.56\linewidth}@{}}
\toprule
Test & Ran on & Rejected & What was rejected \\
\midrule
\multicolumn{4}{@{}l}{\emph{Not triggered in this batch}} \\
Encoding & 69 & 0 & NA \\
Rule & 68 & 0 & NA \\
Outcome & 69 & 0 & NA \\
Information hiding & 43 & 0 & NA \\
\midrule
\multicolumn{4}{@{}l}{\emph{Triggered}} \\
Playability & 33 & 1 & a negotiation family: $0.23$ trades per episode (threshold $0.5$), decisive in $0.20$ of episodes (threshold $0.25$) \\
Agreement & 42 & 1 & a deduction family: $30$ actions against $31$ in the re-implementation, reviewed as an ambiguity in the specification \\
Duplicate (pairs) & 35 & 3 & three families whose outcomes matched an admitted family on all $60$ shared seeds \\
\midrule
\multicolumn{4}{@{}l}{\emph{Checks outside the seven tests}} \\
Static safety & 80 & 4 & proposals using a banned identifier, never reached the replay tests \\
Degeneracy flag & 61 & 1 & a family with a first-mover win share of $0.054$, outside the admissible band \\
\bottomrule
\end{tabular}
\label{tab:gate-ledger}
\end{table}

\paragraph{Injected defects.}
Because most tests were never triggered, we measure their power by calibration. Each of the three seed games is first run through the whole battery unmodified, and every applicable test passes. We then inject eight classes of known defect, twelve instances in total, into the hooks of every seed game to which they apply, which yields $30$ mutants, and run every test again. Since all variants of a family execute the same hook code, a defect in a hook is shared by every encoding, rule, and size variant of the family. We therefore inject each defect into every variant of the family rather than into a single variant, as injecting into one variant only would make the Encoding test appear to catch every defect and would measure the injection rather than the test. The one exception is the last row of Table~\ref{tab:detect}, which is scoped to a single encoding variant because this is precisely the situation that the Encoding test tests for. The Duplicate test is a predicate over pairs of games rather than over a single game, and is therefore reported separately below.

\begin{table}[h]
\centering\footnotesize
\setlength{\tabcolsep}{4pt}
\caption{Injected defects on the three seed games and the tests that rejected them. Each cell counts the mutants of the row that the test rejected, so one mutant may appear in several columns. Columns follow Table~\ref{tab:certs}.}
\begin{tabular}{@{}L{0.36\linewidth} r rrrrrr r@{}}
\toprule
Injected defect & $n$ & Enc & Rule & Out & Hide & Play & Agr & Caught \\
\midrule
\multicolumn{9}{@{}l}{\emph{Outcome and reward hooks}} \\
Verdict inverted & 3 & 0 & 0 & 0 & 0 & 0 & 3 & 3/3 \\
Rewards not zero-sum & 6 & 0 & 3 & 6 & 0 & 1 & 6 & 6/6 \\
Never terminates & 3 & 0 & 1 & 3 & 0 & 1 & 3 & 3/3 \\
\midrule
\multicolumn{9}{@{}l}{\emph{Action rendering}} \\
Action labels corrupted & 6 & 0 & 3 & 6 & 0 & 1 & 6 & 6/6 \\
\midrule
\multicolumn{9}{@{}l}{\emph{Hidden state}} \\
Opponent's private state leaked & 4 & 0 & 0 & 0 & 4 & 0 & 0 & 4/4 \\
\midrule
\multicolumn{9}{@{}l}{\emph{Variant-level defects}} \\
Rule variant has no effect & 3 & 0 & 3 & 0 & 0 & 0 & 0 & 3/3 \\
Menu not at parity, bot never trades & 2 & 0 & 1 & 0 & 0 & 2 & 2 & 2/2 \\
One encoding settles differently & 3 & 3 & 0 & 0 & 0 & 0 & 0 & 3/3 \\
\midrule
Total & 30 & 3 & 11 & 15 & 4 & 5 & 20 & 30/30 \\
\bottomrule
\end{tabular}
\label{tab:detect}
\end{table}

The battery rejects all $30$ injected mutants. Three results in Table~\ref{tab:detect} deserve attention.

First, an inverted verdict is caught only by the agreement check. When the hook that decides the winner is inverted throughout a family, rewards still sum to zero, illegal actions are still penalised, every rendered label is still accepted, every episode still terminates, the encoding variants still agree with one another, and an outcome-reversing rule variant is still an exact flip of its parent. Consequently, every mechanical replay test passes, and only the comparison against an independently written implementation of the same specification detects the defect. This is the strongest evidence that the agreement check is effective, and it justifies the cost of the check even though it turned away only one candidate.

Second, three tests are the sole detector of a defect class: Rule for a rule variant with no effect, Encoding for an encoding that settles differently, and Information hiding for a leak, which no other test notices because a leak changes what a player sees without changing any trajectory. The two menu defects reproduce the failure mode that motivated the Playability test: pricing the trade menu by quantity rather than by base value means that any swap that helps the proposer must cost the responder under their own scoring, so a score-maximising bot never trades. Both mutants fail Playability with $0.00$ trades per episode and are also caught by the Agreement test, so Playability provides a direct check of this failure mode rather than the only one.

Third, the Duplicate test behaves as specified when exercised directly. A planted clone of an admitted game matches on $60$ of $60$ shared seeds and is flagged, two genuinely different games on the same board are not, and the one pair flagged in the ledger reproduces at $60$ of $60$ on the admitted configuration but differs on $11$ of $60$ once the deck is enlarged, which shows that the collision is a property of the configuration rather than of the two specifications.

\paragraph{Dependence on the seed set and the driver.}
To check this dependence, we repeated the whole injection matrix with the seed set shifted and the driver's random stream shifted independently, which changes both the seeds of Table~\ref{tab:certs} and the action sequences drawn on them. Every cell of Table~\ref{tab:detect} is unchanged, and the three seed games still pass every applicable test unmodified. None of the conclusions in this subsection therefore depends on the particular seeds of Table~\ref{tab:certs}.

\paragraph{Manual review.}
The agreement check compares two readings of a specification and cannot detect a misreading that both share. To bound this, we read the specification and the hook implementation of ten admitted games side by side, four grid games (Chop-3 Scorers, Clampdown, Pivot Flips, Trap-Go Threshold), four betting games (Modular Sum Showdown, Bitwise XOR High, Target Sum, Median Dominance), and two negotiation games (K-of-Each Set Collection, Tiered Diminishing Returns), and tested every clause whose reading was in doubt on constructed positions. Eight implement their specification exactly. Trap-Go Threshold adds a termination clause that the specification leaves open, ending the game when the player to move has no legal move, and its rules text states this clause. Tiered Diminishing Returns rounds the inventory score to the nearest integer, which the specification does not, and the rounding changes the winner in $12$ of $20{,}000$ random start-and-end pairs. Its rules text gives the unrounded formula, and the score shown to the players at every turn is the rounded value that the engine settles on.

In both cases the value the player is shown at every turn is the value the engine settles on, so the reward stays consistent with what the policy observes, and what the deviation moves is the specification-to-implementation translation rather than the game the model learns. 
Both deviations are also of a kind that a stronger re-implementation model would be more likely to expose, since neither is ambiguous in the specification text; the check's power is limited by the re-implementer rather than by the protocol, and it improves as the model does.

\FloatBarrier

\subsection{Corpus composition}
\label{app:corpus}

\subsubsection{Admitted games}
\label{app:games}
Across all synthesis runs, the designer proposed $80$ games. Of these, $76$ were implemented within the attempt budget and $75$ passed the replay tests, with an average of $1.7$ engineer attempts per game. 
Every one of the \gsNsynth{} admitted games agrees with its independent re-implementation on every replayed outcome. Table~\ref{tab:games} lists the \gsNsynth{} admitted games by template, with a one-line description of the rule that distinguishes each game from its template's seed game. Full rule texts are included in the corpus release. The complexity reported in \S\ref{sec:analysis} is game-tree complexity, $\log_{10}\mathrm{GTC}\approx\bar{d}\log_{10}\bar{b}$, where $\bar{b}$ is the mean branching factor and $\bar{d}$ is the mean game length under random play \citep{allis1994searching}.

\begin{table}[h]
\centering\scriptsize
\setlength{\tabcolsep}{4pt}
\caption{Admitted games by template.}
\begin{tabular}{@{}llL{9.6cm}@{}}
\toprule
Key & Name & Rule in one line \\
\midrule
\multicolumn{3}{@{}l}{\emph{Grid placement (seed: Tic-Tac-Toe)}} \\
ANTICL & Anti-Clump Diagonals & A stone may not touch one's own stones orthogonally. The first diagonal line of $K$ wins, and a player with no legal move loses. \\
CHOP & Chop-3 Scorers & A row or column that reaches a threshold of one's own stones is cleared. The first to complete the target pattern wins. \\
CLAMP & Clampdown & An opponent stone between two of one's own is captured. The first to the target number of captures wins. \\
CROSSL & Crosslink & One side connects north to south and the other west to east. The first connection wins. \\
FIVMIS & Cross-Exception Mis\`ere & Both players place the same mark. Completing a straight line of $K$ loses. \\
MAJSEG & Majority Segments & The board is cut into segments of $K$ cells. The player holding the majority in more segments wins. \\
NEUTRW & Neutral Win, Hot Loss & Each turn places a neutral or a hot stone. A line of neutral stones wins and a line of hot stones loses. \\
NOSQRE & No-Squares Mis\`ere & Completing a one-colour $k\times k$ square loses. \\
PIVOT & Pivot Flips & A placed stone flips the stones on its orthogonal lines. The first line of $K$ wins. \\
QUARN & Quarantine & The board is split into quadrants and the winning condition must be met inside the designated one. \\
SWEEPR & Row-Column Sweeper & A placement clears the rest of its row and column. The first to complete the target pattern wins. \\
TRAPGO & Trap-Go Threshold & Connected opponent groups whose liberties fall below a threshold are captured. \\
TWINRS & Twin Rows & Lines count only as twin pairs. The first to collect $R$ pairs wins. \\
\midrule
\multicolumn{3}{@{}l}{\emph{Betting with a showdown (seed: Kuhn Poker)}} \\
BITXOR & Bitwise XOR High & The showdown is won by the higher bitwise XOR of the hand's ranks. \\
CENTER & Center-Seek High & The showdown is won by the card closest to a target value. \\
CLUST & Clustered High & The showdown is won by the hand whose ranks are closest together. \\
DUPES & Duplicate Power Bonus & Repeated ranks score as a power of their count, and the higher total wins the showdown. \\
MAXGAP & Far Apart Wins & The showdown is won by the larger gap between the highest and lowest card. \\
MEDIAN & Median Dominance & The showdown is won by the higher median rank. \\
MODSUM & Modular Sum Showdown & The showdown is won by the higher sum of ranks modulo $M$. \\
PARITY & Parity Majority High & The showdown is won by the hand with more cards of the preferred parity, then by the higher card. \\
TARGET & Target Sum, Don't Bust & The showdown is won by the rank sum closest to a target without exceeding it. \\
\midrule
\multicolumn{3}{@{}l}{\emph{Resource negotiation (seed: Simple Negotiation)}} \\
CAPWST & Caps and Waste & Units of a resource above a cap are wasted. The larger growth in inventory value wins. \\
PAIRED & Paired Complements & Resources held in pairs earn a bonus. The larger growth in inventory value wins. \\
SETK & K-of-Each Set Collection & Each complete set of $K$ of every resource earns a bonus. The larger growth in inventory value wins. \\
TIERD & Tiered Diminishing Returns & A resource's value diminishes by tier. The larger growth in inventory value wins. \\
\bottomrule
\end{tabular}
\label{tab:games}
\end{table}

\subsubsection{Training, validation, and held-out variants}
\label{app:curation}
The effect size of a rule or size variant is defined as the fraction of $60$ seed-paired episodes whose outcome differs from that of the parent game under the same action sequence. 
All \gsNfamilies{} families are used for distillation, and $22$ of the synthesised families are sampled by the game sampler during the self-play RL stage.
Within a family, the distillation data consist of the base game with a random encoding per episode, up to two rule variants with effect size in $[0.2, 0.95]$, and the first size variant. Two fixed encodings form a validation split, while the transposed encoding, the rule variant with the highest effect size, and the last size variant are held out as a test split and are not trained on.

\section{Proofs and additional theoretical analysis}
\label{app:theory}

\subsection{Proof of Lemma~\ref{prop:variance}}
\label{app:proof}
We write $S_i=\nabla_\theta\log\pi_\theta(\tau_i)$ and $\hat{g}_i=(R(\tau_i)-b_i)\,S_i$. Since the baseline $b_i$ is a function of the other returns of the batch and of a prior fitted on earlier batches, it is independent of $\tau_i$ given the policy, and $\mathbb{E}[\hat{g}_i]=\mathbb{E}[R(\tau_i)S_i]-\mathbb{E}[b_i]\,\mathbb{E}[S_i]=\nabla_\theta J$ because $\mathbb{E}[S_i]=0$. For the variance,
\begin{equation*}
\mathrm{Var}[\hat{g}_i]\;=\;\mathbb{E}\big[(R(\tau_i)-b_i)^2\,\|S_i\|^2\big]-\|\nabla_\theta J\|^2\;=\;\mathbb{E}\big[(R(\tau_i)-b_i)^2\big]\,\mathbb{E}\big[\|S_i\|^2\big]-\|\nabla_\theta J\|^2,
\end{equation*}
where the second equality follows from the stated assumption. Expanding about $\mu_g$,
\begin{equation*}
\mathbb{E}\big[(R(\tau_i)-b_i)^2\big]\;=\;\mathbb{E}\big[(R(\tau_i)-\mu_g)^2\big]+\mathbb{E}\big[(b_i-\mu_g)^2\big]-2\,\mathbb{E}\big[(R(\tau_i)-\mu_g)(b_i-\mu_g)\big],
\end{equation*}
where the cross term vanishes because $b_i$ is independent of $R(\tau_i)$ and $\mathbb{E}[R(\tau_i)-\mu_g]=0$. Hence, $\mathrm{Var}[\hat{g}_i]=\big(\sigma^2+\mathbb{E}[(b_i-\mu_g)^2]\big)\,\mathbb{E}[\|S_i\|^2]-\|\nabla_\theta J\|^2$, which is an increasing affine function of the mean squared error of $b_i$.

For the second claim, we write $w=K_g'/(K_g'+\kappa)$ so that $b_i-\mu_g=w\,(\bar{y}_{g,-i}-\mu_g)+(1-w)\,(m_g-\mu_g)$. The first term is the sampling error of the leave-one-out mean, with variance $\sigma^2/K_g'$, and the second term is the error of the prior, with mean square $\tau^2$ by definition. The two terms are uncorrelated, since the prior is fixed before the batch is sampled, so
\begin{equation*}
\mathbb{E}\big[(b_i-\mu_g)^2\big]\;=\;w^2\,\frac{\sigma^2}{K_g'}+(1-w)^2\,\tau^2 .
\end{equation*}
This is a convex quadratic in $w$, which is minimised at $w^\star=\tau^2/(\tau^2+\sigma^2/K_g')=K_g'/(K_g'+\sigma^2/\tau^2)$, that is, at $\kappa^\star=\sigma^2/\tau^2$, where its value is $\sigma^2\tau^2/(K_g'\tau^2+\sigma^2)=\sigma^2/(K_g'+\kappa^\star)$. The two limiting cases are $w=1$, the leave-one-out mean with error $\sigma^2/K_g'$, and $w=0$, the prior with error $\tau^2=\sigma^2/\kappa^\star$, and both exceed $\sigma^2/(K_g'+\kappa^\star)$.

\subsection{Error analysis of the shrinkage baseline}
\label{app:error}
The proof above gives the mean squared error of the baseline in closed form,
\begin{equation}
\label{eq:mse}
\mathrm{MSE}(w)\;=\;\mathbb{E}\big[(b_i-\mu_g)^2\big]\;=\;w^2\,\frac{\sigma^2}{K_g'}+(1-w)^2\,\tau^2,\qquad w=\frac{K_g'}{K_g'+\kappa},
\end{equation}
and three consequences of this expression are used in the paper.

First, the decomposition does not assume that the prior is correct. Since $\tau^2$ is defined as the mean squared error of the prior that is actually used, it includes both the bias of the linear model, which cannot represent interactions between fields, and its lag, since the prior is fitted on earlier batches. The normal model of \S\ref{sec:method:ourrl} is used only to motivate the choice $\kappa^\star=\sigma^2/\tau^2$. For any fixed $\kappa$, the baseline is independent of the action, the gradient remains unbiased, and \eqref{eq:mse} gives its error.

Second, the error is flat near its minimum. Since \eqref{eq:mse} is a quadratic in $w$ with second derivative $2(\sigma^2/K_g'+\tau^2)$,
\begin{equation}
\label{eq:mse-flat}
\mathrm{MSE}(w)-\mathrm{MSE}(w^\star)\;=\;(w-w^\star)^2\,\Big(\frac{\sigma^2}{K_g'}+\tau^2\Big).
\end{equation}
This observation has two uses. On the one hand, a fixed $\kappa$ that is within a factor of two of $\kappa^\star$ costs little, which is why the main runs fix $\kappa=4$. On the other hand, replacing $\kappa^\star$ by a plug-in estimate $\hat{\kappa}$ moves $w$ by some $\delta$ and costs $\delta^2(\sigma^2/K_g'+\tau^2)$, so the estimated variant of the estimator is not expected to differ from the fixed one unless $\hat{\kappa}$ is far off.

The quantities in \eqref{eq:mse} are estimated during training. The within-group variance $\sigma^2$ is estimated by pooling the sample variances of the groups of a batch, and $\tau^2=\mathbb{E}[(\mu_g-m_g)^2]$ is estimated by the method of moments,
\begin{equation}
\label{eq:tau}
\hat{\tau}^2\;=\;\operatorname{mean}_g\Big[\big(\bar{y}_g-m_g\big)^2-\hat{\sigma}^2/K_g\Big],
\end{equation}
where $\bar{y}_g$ is the full group mean and the second term removes its sampling noise. Both estimates are smoothed across batches, and $\hat{\kappa}=\hat{\sigma}^2/\hat{\tau}^2$. The variant of \ourrl{} that uses $\hat{\kappa}$ does so with one batch of lag, so that the baseline of a batch never depends on its own returns. Because $\tau^2$ is measured against the prior that is actually used, it includes the prior's bias and lag. The encoding variance $\hat{\tau}^2_{\mathrm{enc}}$ is computed on the same returns split by encoding, as the variance across encodings of the mean return within each group, corrected for sampling noise in the same way.

Third, since these quantities are measured, the error can be evaluated empirically. Table~\ref{tab:eb} reports $\hat{\sigma}^2$, $\hat{\tau}^2$ of \eqref{eq:tau}, $\hat{\kappa}=\hat{\sigma}^2/\hat{\tau}^2$, and $\hat{\tau}^2_{\mathrm{enc}}$ on Qwen3-4B-Base with $G=5$, averaged over steps $31$ to $60$ of the main configuration with $\kappa=4$ and of the variant that uses $\hat{\kappa}$ with one batch of lag. With $K_g'=4$, the leave-one-out mean alone has error $\sigma^2/K_g'\approx0.23$, the prior alone has error $\tau^2\approx0.16$, and the shrinkage estimator has error $0.10$ at either $\kappa=4$ or $\kappa=\hat{\kappa}$. The estimate $\hat{\kappa}$ lies between $5$ and $8$ from step $10$ onwards, so $\kappa=4$ is within a factor of two of the estimated optimum, and by \eqref{eq:mse-flat} the difference in error between the two runs is below $0.003$. Their scores are consistent with this: at steps $20$, $40$, and $60$ they differ by $-1.6$, $-1.7$, and $+0.1$ points, with no consistent sign. The encoding variance $\hat{\tau}^2_{\mathrm{enc}}$ decreases from $0.70$ at step $15$ to $0.50$ at step $30$ and stays there, indicating that the policy moves towards encoding invariance early in training but remains some distance from it at this budget. Note that these are moment estimates from the returns of each batch rather than held-out prediction errors, and that the two runs are single seeds.

\begin{table}[h]
\centering\small
\caption{Estimated variances of the shrinkage baseline on Qwen3-4B-Base, steps $31$ to $60$, and the baseline error implied by \eqref{eq:mse}.}
\begin{tabular}{@{}lccccc@{}}
\toprule
Run & $\hat{\sigma}^2$ & $\hat{\tau}^2$ & $\hat{\kappa}$ & $\hat{\tau}^2_{\mathrm{enc}}$ & $\mathrm{MSE}$ at $\kappa$ used \\
\midrule
$\kappa=4$ & 0.91 & 0.16 & 6.4 & 0.50 & 0.098 \\
$\kappa=\hat{\kappa}$ & 0.94 & 0.19 & 5.4 & 0.49 & 0.099 \\
\bottomrule
\end{tabular}
\label{tab:eb}
\end{table}

Fourth, the error can be measured directly rather than inferred from \eqref{eq:mse}. The claim under test is that of Lemma~\ref{prop:variance}: the shrinkage baseline is an interior point between the leave-one-out mean at $w=1$ and the prior at $w=0$, its error is smaller than that of either endpoint, and \eqref{eq:mse} gives its size. To measure this, we freeze the step-$40$ and step-$120$ checkpoints of the run with $\kappa=4$ and draw independent rollout batches from each with the training sampler and the training hyperparameters ($G=5$, $K_g'=4$, temperature $1$, response budget $16{,}384$ tokens), with no parameter update: $35$ and $96$ batches of the size that the trainer hands to one lane, i.e., $535$ and $1{,}500$ games for the two steps, respectively. Since $\mu_g$ is fixed, every baseline can be recomputed offline on the same returns, so the differences between rows come from the baseline alone. Because $b_i$ never uses $R_i$, $\mathbb{E}[(R_i-b_i)^2]=\sigma^2+\mathbb{E}[(b_i-\mu_g)^2]$, and subtracting the pooled within-group variance ($\hat{\sigma}^2=0.95$ and $0.92$ over the whole probe) gives the error of each baseline without a proxy for $\mu_g$. Two baselines are compared as a paired difference of $(R_i-b_i)^2$ on the same trajectories, which removes the variation they share and does not depend on $\hat{\sigma}^2$. Intervals are $95\%$ bootstrap intervals over (batch, group) blocks. The factored prior is fitted online, lane by lane, on the earlier batches of the probe, as in training, with the first $20$ batches excluded as warm-up. The closed form \eqref{eq:mse} is evaluated at the $\hat{\sigma}^2$ above and at $\hat{\tau}^2=0.15$ and $0.12$ from \eqref{eq:tau} on the same rollouts, which give $\hat{\kappa}=6.3$ and $7.4$. Table~\ref{tab:eb-measured} reports the measured errors and Table~\ref{tab:eb-kappa} sweeps $\kappa$ on the same rollouts. Three findings support the claim.

The interior point beats both endpoints. At step $120$ the leave-one-out mean has a measured error of $0.240$, the prior $0.129$, and \ourrl{} with $\kappa=4$ $0.096$. The paired differences agree, with intervals that exclude zero: \ourrl{} is $0.143$ $[0.129, 0.157]$ below the leave-one-out mean and $0.033$ $[0.006, 0.063]$ below the prior. Step $40$ gives the same ordering, $0.233$, $0.148$, and $0.093$, again with paired intervals that exclude zero.

The closed form predicts the measured error. Evaluated at the $\hat{\sigma}^2$ and $\hat{\tau}^2$ of the same rollouts, \eqref{eq:mse} is within $0.010$ of the measurement on every row of Table~\ref{tab:eb-measured}, and the measured error of the estimator used in the main runs, $0.093$ and $0.096$, agrees with the $0.098$ that Table~\ref{tab:eb} infers from \eqref{eq:mse} during training. The probe also reproduces the moment estimates of the training run, $\hat{\sigma}^2=0.95$ and $0.92$ against $0.91$, and $\hat{\kappa}=6.3$ and $7.4$ against $6.4$, which serves as a check on the probe itself.

The fixed $\kappa$ costs little. Table~\ref{tab:eb-kappa} sweeps $\kappa$ on the same rollouts. The measured curve is a shallow basin with its minimum at $\kappa=8$, next to the $\hat{\kappa}$ of these rollouts, and it is flat across the range that \eqref{eq:mse-flat} predicts to be flat: moving from $\kappa=4$ to the minimum changes the error by $0.003$ at step $40$ and by $0.009$ at step $120$, compared with a gap of about $0.14$ to the leave-one-out mean.

The lemma bears on the error of the baseline, which is what these tables measure, and the comparison between estimators in training is the empirical one of Table~\ref{tab:pipelines}.

\begin{table}[h]
\centering\footnotesize
\setlength{\tabcolsep}{3pt}
\caption{Measured error of the baseline under a frozen policy, at the step-$40$ and step-$120$ checkpoints of the \ourrl{} run of Table~\ref{tab:eb}. Measured MSE is $\mathbb{E}[(R_i-b_i)^2]-\hat{\sigma}^2$, the \eqref{eq:mse} column is the closed form at the $\hat{\sigma}^2$ and $\hat{\tau}^2$ of the same rollouts, and $\Delta$ is the paired difference against \ourrl{} with $\kappa=4$, positive when the row has the larger error, with $95\%$ bootstrap intervals in brackets.}
\begin{tabular}{@{}rlclcl@{}}
\toprule
Step & Baseline & $w$ & Measured MSE & \eqref{eq:mse} & $\Delta$ vs \ourrl{}, $\kappa=4$ \\
\midrule
40 & \ourrl{}, $\kappa=4$ & 0.50 & 0.093 [0.044, 0.142] & 0.097 & -- \\
 & \ourrl{}, $\kappa=\hat{\kappa}$ & 0.39 & 0.089 [0.043, 0.133] & 0.092 & $-$0.005 [$-$0.013, 0.004] \\
 & Leave-one-out mean & 1 & 0.233 [0.166, 0.300] & 0.238 & 0.139 [0.115, 0.163] \\
 & Factored prior & 0 & 0.148 [0.101, 0.197] & 0.151 & 0.055 [0.006, 0.106] \\
\midrule
120 & \ourrl{}, $\kappa=4$ & 0.50 & 0.096 [0.065, 0.127] & 0.088 & -- \\
 & \ourrl{}, $\kappa=\hat{\kappa}$ & 0.35 & 0.087 [0.058, 0.115] & 0.080 & $-$0.009 [$-$0.015, $-$0.001] \\
 & Leave-one-out mean & 1 & 0.240 [0.195, 0.281] & 0.230 & 0.143 [0.129, 0.157] \\
 & Factored prior & 0 & 0.129 [0.096, 0.160] & 0.124 & 0.033 [0.006, 0.063] \\
\bottomrule
\end{tabular}
\label{tab:eb-measured}
\end{table}

\begin{table}[h]
\centering\footnotesize
\setlength{\tabcolsep}{4pt}
\caption{The rollouts of Table~\ref{tab:eb-measured} with $\kappa$ swept. The first and last rows are the two endpoints of Lemma~\ref{prop:variance}, and the \eqref{eq:mse} columns evaluate the closed form at the $\hat{\sigma}^2$ and $\hat{\tau}^2$ of the same rollouts ($\hat{\kappa}=6.3$ and $7.4$).}
\begin{tabular}{@{}lccccc@{}}
\toprule
$\kappa$ & $w$ & Step 40, measured & \eqref{eq:mse} & Step 120, measured & \eqref{eq:mse} \\
\midrule
$0$ (leave-one-out) & 1 & 0.233 [0.166, 0.300] & 0.238 & 0.240 [0.195, 0.281] & 0.230 \\
\midrule
$0.5$ & 0.89 & 0.185 [0.121, 0.248] & 0.190 & 0.192 [0.150, 0.232] & 0.183 \\
$1$ & 0.80 & 0.154 [0.093, 0.214] & 0.158 & 0.161 [0.121, 0.199] & 0.152 \\
$2$ & 0.67 & 0.118 [0.063, 0.173] & 0.122 & 0.124 [0.088, 0.159] & 0.116 \\
$4$ & 0.50 & 0.093 [0.044, 0.142] & 0.097 & 0.096 [0.065, 0.127] & 0.088 \\
$8$ & 0.33 & 0.090 [0.047, 0.131] & 0.094 & 0.087 [0.059, 0.115] & 0.081 \\
$16$ & 0.20 & 0.103 [0.061, 0.143] & 0.106 & 0.095 [0.066, 0.121] & 0.088 \\
\midrule
$\infty$ (prior) & 0 & 0.148 [0.101, 0.197] & 0.151 & 0.129 [0.096, 0.160] & 0.124 \\
\bottomrule
\end{tabular}
\label{tab:eb-kappa}
\end{table}

\subsection{Encoding symmetry}
\label{app:symmetry}
Fix a game and let $\mathcal{G}$ be its set of encodings. Each $e\in\mathcal{G}$ is a bijection on rendered observations and on action labels, and the kernel of the template applies it after the game logic has run (\S\ref{sec:method:templates}): the state is stored in a canonical form, the hooks act on that form, and $e$ renders the state and reads the action. Since the composition and inversion of such maps are again such maps, $\mathcal{G}$ is a group, and for every state $s$ and action $a$ the canonical successor and outcome of $(s,a)$ are those of $(e(s),e(a))$. Three statements follow from this.

First, values are invariant. Since $V$ is computed from the canonical state by the hooks (\S\ref{sec:method:rl}), we have $V(e(s))=V(s)$. Similarly, the minimax value of the game from a seat is a function of the canonical game tree, which $e$ does not change.

Second, the group maps optimal policies to optimal policies. For a policy $\pi$ and $e\in\mathcal{G}$, define $\pi^{e}(e(a)\mid e(s))=\pi(a\mid s)$. By the commutation above, the distribution over canonical trajectories induced by $\pi^{e}$ playing under encoding $e$ equals the distribution induced by $\pi$ playing under the identity, for either seat and against any opponent transported in the same way. Since expected returns are functions of canonical trajectories, they agree, and a policy that is optimal under the identity is therefore transported to a policy that is optimal under $e$.

Third, equivariant policies have encoding-independent returns. If $\pi^{e}=\pi$ for every $e$, then playing $\pi$ under $e$ is the same as playing $\pi^{e}$ under $e$, which by the previous statement has the expected return of $\pi$ under the identity. Hence, the expected return of a group $g$ does not depend on the encoding, the variance across encodings of the mean return within a group is zero, and $\tau^2_{\mathrm{enc}}$ of \S\ref{sec:method:ourrl} vanishes. Conversely, a nonzero $\tau^2_{\mathrm{enc}}$ indicates that the policy treats some renderings of the same game differently.

\subsection{Equivariance and the encoding variants}
\label{app:encoding-efficiency}
The encoding group of Appendix~\ref{app:symmetry} also characterises what training under a fresh encoding per episode aims at, and why it does not enlarge what has to be learned. Fix a game with encoding group $\mathcal{G}$, and for a policy $\pi$ write $\pi^{e}$ for its transport by $e\in\mathcal{G}$ as above. The \emph{orbit average} of $\pi$ is the mixture
\begin{equation*}
Q\pi\;=\;\frac{1}{|\mathcal{G}|}\sum_{e\in\mathcal{G}}\pi^{e},
\end{equation*}
which is itself a policy, and since $e'\in\mathcal{G}$ permutes the terms of the sum, $(Q\pi)^{e'}=Q\pi$. Therefore, $Q$ maps every policy to an equivariant one and fixes the equivariant ones, that is, $Q$ is a projection onto the equivariant policies. Let $L_e(\pi)$ be the distillation loss of \S\ref{sec:method:distill} on teacher trajectories rendered under encoding $e$, i.e., the cross-entropy of $\pi$ against the action of $\pi_{\mathrm{T}}$. Drawing the encoding uniformly per episode, as the distillation data do, minimises the orbit average of the loss, $\bar{L}(\pi)=|\mathcal{G}|^{-1}\sum_{e}L_e(\pi)$. The argument assumes that the teacher is equivariant, so that its trajectories and actions are canonical objects rendered under $e$. Three facts follow.

First, $\bar{L}$ is invariant under transport. By the commutation of Appendix~\ref{app:symmetry}, $\pi^{e'}$ evaluated on trajectories rendered under $e$ equals $\pi$ evaluated on the same trajectories rendered under $e'^{-1}\circ e$, so $L_e(\pi^{e'})=L_{e'^{-1}\circ e}(\pi)$, and averaging over $e$ gives $\bar{L}(\pi^{e'})=\bar{L}(\pi)$.

Second, the orbit average does not increase the loss. Since the cross-entropy is convex in the policy's conditional distributions, Jensen's inequality together with the first fact gives $\bar{L}(Q\pi)\le|\mathcal{G}|^{-1}\sum_{e}\bar{L}(\pi^{e})=\bar{L}(\pi)$. A minimiser of $\bar{L}$ can therefore be taken to be equivariant, and training under random encodings targets an equivariant policy, which by Appendix~\ref{app:symmetry} is also where the optimal self-play policy lies.

Third, the equivariant class is no larger than the full class. Let $\Pi$ be the set of policies that the model can represent, and measure distance by $d(\pi,\pi')=\max_{s}\sum_{a}|\pi(a\mid s)-\pi'(a\mid s)|$. Since transport by $e$ is a relabelling of states and actions, we have $d(\pi^{e},\pi'^{e})=d(\pi,\pi')$, and the triangle inequality gives $d(Q\pi,Q\pi')\le d(\pi,\pi')$. The image under $Q$ of an $r$-cover of $\Pi$ is an $r$-cover of $Q\Pi$, so the covering number of $Q\Pi$ is at most that of $\Pi$, and the covering-number bounds on the generalisation gap \citep{zhou2002covering,bartlett2002rademacher} for the equivariant class are no larger than those for $\Pi$. This is the same argument that \citet{qin2022permutation} make for permutation symmetry in auctions and \citet{vanderknaap2026prism} for reflection symmetry in control, applied here to the encoding group, which is finite and known exactly for every game because the kernel implements it.

Two limitations of this argument are worth stating. First, the policy is not constrained to $Q\Pi$: an LLM reads rendered text and is not equivariant by construction, so random encodings change the target of training rather than the class, and $\hat{\tau}^2_{\mathrm{enc}}$ of Appendix~\ref{app:error} measures how far the trained policy still is from that target. This is also why encodings are needed as training variants at all. For an equivariant model, one rendering of a game would be as good as all of them, but a model that has to learn the invariance learns it from the renderings it sees, and these are the cheapest variants in the corpus, since they add no game to admit and no reward to verify. The second limitation is quantitative. A reduction in sample complexity by a factor related to $|\mathcal{G}|$ is known for linear and kernel models \citep{elesedy2021provably} and under the general framework of \citet{chen2020group}, but it is not claimed here for a transformer policy. What the argument does establish is that the target reached by training on random encodings is the same game-theoretic policy in every rendering, that this target is no harder to represent than the policy for one rendering, and that once the policy is equivariant, a trajectory under any encoding carries the same information about every other. This is the sense in which the corpus counts encoding variants separately from rule and size variants.

\section{Additional implementation details}
\label{app:protocol}

\subsection{Non-game control data}
\label{app:teacher-data}
The \emph{non-game} control (Table~\ref{tab:general-reasoning}) isolates whether the transfer is due to distillation of a strong teacher's reasoning rather than to something specific to games. It uses the \emph{same} teacher, decoding, and $\langle$think$\rangle$\,$+$\,\texttt{\textbackslash boxed} output format as the \gamesmith{} SFT data, but on non-game reasoning prompts spanning the capability types that the games exercise. Prompts are drawn only from the \emph{training} splits of the sources in Table~\ref{tab:teacher-data} and sampled with the teacher, and only teacher solutions that are verified correct against the gold answer are kept. The retained set is subsampled to the \gamesmith{}~(matched) token budget.

\begin{table}[t]
\centering\small
\caption{Non-game control data: verified-correct teacher traces on non-game reasoning, by capability and source (training splits only), decontaminated against every evaluation benchmark and subsampled to the \gamesmith{} (matched) token budget.}
\begin{tabular}{@{}llr@{}}
\toprule
Capability & Source (train split) & Traces \\
\midrule
Deductive logic       & ReClor       & 7{,}666 \\
                                       & LogiQA       & 6{,}163 \\
                                       & ProofWriter  & 2{,}334 \\
\midrule
Spatial \& state-tracking & SpartQA  & 6{,}339 \\
                                       & StepGame     & 184 \\
\midrule
Competition mathematics & MetaMathQA        & 7{,}889 \\
                                       & OpenMathInstruct-2  & 5{,}777 \\
                                       & NuminaMath-CoT      & 3{,}690 \\
\midrule
\textbf{Total} & & \textbf{40{,}042} \\
\bottomrule
\end{tabular}
\label{tab:teacher-data}
\end{table}

Before training, we remove any prompt that overlaps with the evaluation data, checked against \emph{every} benchmark in Table~\ref{tab:general-reasoning} (the five mathematics sets, MMLU-Pro, GPQA-Diamond, and the six BBH tasks). A prompt is dropped if it triggers any of three independent checks: (i) exact match after normalisation, (ii) a shared $13$-gram with an evaluation question, or (iii) an embedding near-duplicate, with cosine similarity $\ge 0.85$ to the nearest evaluation question under a sentence encoder. Of $26{,}904$ candidate prompts, $111$ were dropped ($14$ exact, $67$ by $13$-gram, $30$ by embedding). These were concentrated in the mathematics sources, whose items are frequently rephrasings of MATH/AMC/GSM8K, whereas the logical and spatial sources triggered essentially no overlap. We err on the side of dropping.

\subsection{Hyperparameters, software, and hardware}
\label{app:hparams}
Table~\ref{tab:hparams} lists the settings of the three stages. Every distillation run is trained with the same optimiser settings for a fixed number of optimiser steps ($2{,}056$ for the settings of Table~\ref{tab:general-reasoning}, $2{,}048$ for the SPIRAL setting) and is evaluated at its final weights. No checkpoint selection is applied to any distillation run, including those of Figure~\ref{fig:rq3-envscaling}, and intermediate checkpoints are kept only for crash recovery. 

\begin{table}[h]
\centering\footnotesize
\setlength{\tabcolsep}{4pt}
\renewcommand{\arraystretch}{1.15}
\caption{Hyperparameters, software, and hardware of the three stages.}
\begin{tabular}{@{}llL{7.6cm}@{}}
\toprule
Stage & Setting & Value \\
\midrule
Distillation (SFT)
 & optimiser, learning rate & AdamW, $10^{-5}$, cosine to zero, warmup $3\%$ \\
 & effective batch & $32$ sequences, per-device batch $1$ with gradient accumulation \\
 & context & $16{,}384$ tokens, no truncation of any record \\
 & precision, memory & bf16, gradient checkpointing, fused cross-entropy \\
 & parallelism & data parallel for 4B, ZeRO-2 for 8B and 12B \\
 & GPUs & $4$ GPUs for 4B and 8B, $8$ GPUs for 12B \\
\midrule
Self-play (RL)
 & estimator & \ourrl{} with $\kappa=4$, $\lambda=0.5$, and $G=5$ episodes per sampled variant, each with a fresh seed and encoding. The RAE control on the \gamesmith{} corpus keeps $G=5$ and every other setting and replaces only the estimator. \\
 & optimiser & learning rate $10^{-6}$, PPO clip ratio $0.2$, no KL or entropy term \\
 & sampling & temperature $1.0$, top-$p$ $1.0$, at most $50$ turns per episode \\
 & context & prompt up to $4{,}096$ tokens and response up to $16{,}384$ in a $20{,}480$-token window.  \\
 & over-length & a response beyond $15{,}360$ tokens receives $-1.5$, folded into its return before the baseline is subtracted. \\
 & batch & $128$ transitions (decisions) per optimiser step, $200$ steps \\
 & software & verl~0.6.1 with vLLM~0.10.2 rollouts, FSDP actor \\
 & GPUs & $4$ GPUs for the 4B model, $8$ GPUs with tensor parallelism $2$ for the 8B, 9B, and 12B models \\
\midrule
Evaluation
 & decoding & $4$ samples, temperature $0.6$, top-$p$ $0.95$, $8{,}192$ tokens \\
 & software & vLLM~0.10.2, one GPU per checkpoint \\
 & wall-clock & about $5$ hours for the $13$ benchmarks on an 8B checkpoint \\
\bottomrule
\end{tabular}
\label{tab:hparams}
\end{table}

\subsection{Compute budget}
\label{app:compute}

All training and evaluation runs use GPUs with 16,896 CUDA cores, 132 streaming multiprocessors, and 80 GB of memory.
Each run uses a single node with 8 GPUs, with one GPU for evaluation and 4 or 8 GPUs for training, as specified in Table~\ref{tab:hparams}.
Compute is reported in GPU hours, calculated as wall-clock hours multiplied by the number of GPUs.
Training estimates are derived from per-step timings in the logs.
All estimates are rounded to two significant figures.

The main training and evaluation experiments require approximately $9{,}800$ GPU hours.
The twelve self-play runs in Table~\ref{tab:pipelines} account for approximately $9{,}100$ GPU hours at the $200$-step budget.
The sixteen distillation runs in Table~\ref{tab:general-reasoning} require approximately $360$ GPU hours in total, and evaluation of the checkpoint grids of Table~\ref{tab:pipelines} requires a further $380$ GPU hours.
The eight $G=1$ runs of Appendix~\ref{app:group-size} add approximately $4{,}200$ GPU hours, and the game-side evaluation of Appendix~\ref{app:game-strength} about $200$ GPU hours on single GPUs.
In total, the study uses approximately $14{,}000$ GPU hours across all experiments reported in the main paper and appendices.

\subsection{Evaluation protocol}
\label{app:eval}
Table~\ref{tab:benchmarks} lists the $13$ benchmarks, their sizes, and their groups. Every benchmark is evaluated on its full test set, except for MMLU-Pro, for which a fixed random subset of $1{,}000$ questions is used. Each question is sampled $4$ times at temperature $0.6$ and top-$p$ $0.95$ with an $8{,}192$-token generation budget under vLLM. A sample is considered correct if its final \texttt{\textbackslash boxed} answer matches the reference, under a symbolic-equivalence grader for mathematics and exact match after normalisation otherwise. pass@1 is the mean correctness over the $4$ samples, and the headline number is the unweighted mean of the $13$ per-benchmark pass@1 values. The same protocol is applied to every checkpoint in the paper, including the base models, with a context window of $32{,}768$ tokens for distillation checkpoints and $12{,}288$ for self-play checkpoints, both of which are larger than any prompt plus the generation budget.

\begin{table}[ht]
\centering\small
\caption{Evaluation benchmarks, sizes, and groups. The five mathematics sets are the ones distributed with the SPIRAL evaluation code.}
\begin{tabular}{@{}llr@{}}
\toprule
Group & Benchmark & Questions \\
\midrule
Mathematics & AIME 2024 & 30 \\
 & AMC 12 (2022--2023) & 83 \\
 & MATH-500 & 500 \\
 & Minerva Math & 272 \\
 & OlympiadBench & 675 \\
\midrule
Knowledge & MMLU-Pro (fixed random subset) & 1{,}000 \\
 & GPQA-Diamond & 198 \\
\midrule
Logical reasoning & BBH Logical Deduction (3 obj.) & 250 \\
 & BBH Logical Deduction (5 obj.) & 250 \\
 & BBH Web of Lies & 250 \\
\midrule
Spatial \& tracking & BBH Tracking Shuffled Objects (3 obj.) & 250 \\
 & BBH Geometric Shapes & 250 \\
 & BBH Navigate & 250 \\
\bottomrule
\end{tabular}
\label{tab:benchmarks}
\end{table}

\subsection{Assets and licences}
\label{app:licences}

\begin{table}[ht]
\centering\small
\setlength{\tabcolsep}{4pt}
\caption{Assets and licences.}
\begin{tabular}{@{}L{0.64\linewidth} L{0.34\linewidth}@{}}
\toprule
Asset & Licence \\
\midrule
\multicolumn{2}{@{}l}{\emph{Models}} \\
Qwen3-4B-Base, Qwen3-8B-Base & Apache 2.0 \\
Qwen3-32B (teacher, independent re-implementation) & Apache 2.0 \\
Mistral-Nemo-Instruct-2407 & Apache 2.0 \\
GLM-4-9B-0414 & MIT \\
GPT-5 through its API (designer and engineer) & OpenAI terms of use \\
\midrule
\multicolumn{2}{@{}l}{\emph{Software and code}} \\
TextArena (seed games) & MIT \\
SPIRAL (evaluation code, mathematics sets, RAE) & MIT \\
verl, vLLM (training, inference) & Apache 2.0 \\
\midrule
\multicolumn{2}{@{}l}{\emph{Evaluation datasets}} \\
MATH-500, MMLU-Pro, BBH & MIT \\
AIME 2024, AMC 12 (2022--2023), Minerva Math & as distributed with SPIRAL \\
OlympiadBench & MIT \\
GPQA-Diamond & CC BY 4.0 \\
\midrule
\multicolumn{2}{@{}l}{\emph{Non-game control datasets}} \\
ReClor & non-commercial research use \\
LogiQA & CC BY-NC-SA 4.0 \\
ProofWriter, OpenMathInstruct-2 & CC BY 4.0 \\
SpartQA, StepGame, MetaMathQA & MIT \\
NuminaMath-CoT & Apache 2.0 \\
\bottomrule
\end{tabular}
\label{tab:licences}
\end{table}

Table~\ref{tab:licences} lists the models, software, and datasets used in this work, together with their licences. ReClor and LogiQA are licensed for non-commercial research use and are used only to build the non-game control, which is not redistributed. The synthesised corpus contains only game specifications and code produced by the pipeline of \S\ref{sec:method:synthesis}.

\section{Additional empirical results}
\label{app:results}
This appendix supplements \S\ref{sec:experiments} with the diversity and complexity of the corpus, diagnostics of the Format control, and the self-play stage on the games themselves: the evaluation curves of the three pipelines over the budget, the rate of draws and the advantage signal, the comparison of the two estimators with one episode per variant, and the win rates of the trained policy on variants that it never trained on.

\subsection{Corpus diversity and complexity}
\label{app:diversity}
\begin{figure}[h]
\centering
\includegraphics[width=1.0\linewidth]{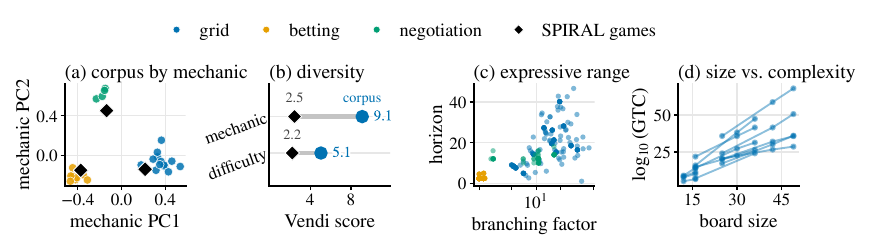}
\caption{Corpus diversity and complexity. \textbf{(a)} Families embedded by rule text (PCA), coloured by genre, with the \gsNseeds{} SPIRAL games shown as black diamonds. \textbf{(b)} Vendi score of the corpus and of the SPIRAL games under a mechanic and a difficulty lens. \textbf{(c)} Branching factor and horizon of all variants and base games. \textbf{(d)} Game-tree complexity against board size, with one line per grid family.}
\label{fig:corpus}
\end{figure}

Figure~\ref{fig:corpus} summarises the diversity and complexity of the corpus. 
Embedding every family by its rule text shows that the synthesised games spread across all \gsNskeletons{} genres, while the \gsNseeds{} SPIRAL games occupy only a small region. 
In terms of effective count, the Vendi score by mechanic is \gsVendiMech{} for the corpus against \gsVendiMechSeeds{} for the three SPIRAL games, and \gsVendiBehav{} against \gsVendiBehavSeeds{} on a behavioural descriptor (decisiveness, first-mover edge, game length, and branching). 
Rule variation changes outcomes: the fraction of episodes whose outcome is flipped by a rule change ranges from \gsEffectMin{} to \gsEffectMax{} (median \gsEffectMedian{}), whereas encoding variants preserve outcomes exactly. 
Game-tree complexity (Appendix~\ref{app:games}) spans $[\gsComplexityMin{},\gsComplexityMax{}]$ across the \gsNvariants{} variants: size variants supply the top of the range, raising the complexity in every grid family and the ceiling from \gsComplexityBaseMax{} to \gsComplexityStructMax{}, betting families keep the game tree small, and negotiation families fill the middle.

\subsection{Per-benchmark results}
\label{app:full-tables}

Tables~\ref{tab:full-qwen34bbase}, \ref{tab:full-qwen38bbase}, \ref{tab:full-mistralnemoinstruct2407}, and \ref{tab:full-glm49b0414} present, for each model of Table~\ref{tab:general-reasoning}, the per-benchmark results that it summarises by capability group.

\begin{table}[htbp]
\centering\footnotesize
\setlength{\tabcolsep}{4pt}
\caption{Per-benchmark results for Qwen3-4B-Base (pass@1 \%, $8{,}192$-token budget). Layout, shading, and bold as in Table~\ref{tab:general-reasoning}.}
\begin{tabular}{@{}l ccccc@{}}
\toprule
 &  & \multicolumn{3}{c}{Controls} & \gamesmith{} \\
\cmidrule(lr){3-5}\cmidrule(lr){6-6}
Benchmark & Base & Format & Non-game & SPIRAL & matched \\
\midrule
\textbf{Mathematics} & 32.5 & 10.7 & 40.6 & 39.6 & \cellcolor{winbg}\textbf{45.7} \\
\quad AIME & 5.0 & 0.0 & 16.7 & 14.2 & \cellcolor{winbg}\textbf{20.8} \\
\quad AMC & 37.0 & 13.0 & 44.0 & 44.0 & \cellcolor{winbg}\textbf{51.8} \\
\quad MATH-500 & 63.2 & 21.1 & 75.5 & 73.4 & \cellcolor{winbg}\textbf{79.4} \\
\quad Minerva Math & 23.9 & 12.2 & 31.2 & 30.8 & \cellcolor{winbg}\textbf{32.8} \\
\quad OlympiadBench & 33.1 & 7.1 & 35.4 & 35.7 & \cellcolor{winbg}\textbf{43.6} \\
\midrule
\textbf{Knowledge} & 42.9 & 34.5 & 47.4 & 48.3 & \cellcolor{winbg}\textbf{49.6} \\
\quad MMLU-Pro & 52.2 & 31.9 & \textbf{62.3} & 59.9 & \cellcolor{winbg}61.1 \\
\quad GPQA-Diamond & 33.6 & 37.0 & 32.4 & 36.7 & \cellcolor{winbg}\textbf{38.0} \\
\midrule
\textbf{Logical reasoning} & 54.9 & 50.9 & 84.1 & \textbf{85.5} & 83.4 \\
\quad BBH Logical Deduction (3 obj.) & 51.1 & 64.0 & \textbf{97.7} & 95.3 & \cellcolor{winbg}95.4 \\
\quad BBH Logical Deduction (5 obj.) & 38.4 & 46.5 & \textbf{83.8} & 81.8 & \cellcolor{winbg}83.4 \\
\quad BBH Web of Lies & 75.1 & 42.1 & 70.7 & \textbf{79.5} & 71.4 \\
\midrule
\textbf{Spatial \& tracking} & 59.5 & 33.1 & \textbf{80.4} & 78.8 & 78.4 \\
\quad BBH Tracking Shuffled Objects & 74.2 & 23.5 & \textbf{96.9} & 90.9 & 87.4 \\
\quad BBH Geometric Shapes & 40.3 & 23.3 & 50.7 & 55.7 & \cellcolor{winbg}\textbf{56.5} \\
\quad BBH Navigate & 64.0 & 52.5 & \textbf{93.6} & 89.7 & \cellcolor{winbg}91.4 \\
\midrule
\textbf{Average (13 benchmarks)} & 45.5 & 28.8 & 60.8 & 60.6 & \cellcolor{winbg}\textbf{62.5} \\
\bottomrule
\end{tabular}
\label{tab:full-qwen34bbase}
\end{table}

\begin{table}[htbp]
\centering\footnotesize
\setlength{\tabcolsep}{4pt}
\caption{Per-benchmark results for Qwen3-8B-Base (pass@1 \%, $8{,}192$-token budget). Layout, shading, and bold as in Table~\ref{tab:general-reasoning}.}
\begin{tabular}{@{}l ccccc@{}}
\toprule
 &  & \multicolumn{3}{c}{Controls} & \gamesmith{} \\
\cmidrule(lr){3-5}\cmidrule(lr){6-6}
Benchmark & Base & Format & Non-game & SPIRAL & matched \\
\midrule
\textbf{Mathematics} & 37.4 & 7.2 & 44.4 & 50.5 & \cellcolor{winbg}\textbf{51.6} \\
\quad AIME & 10.8 & 0.0 & 20.0 & 25.0 & \cellcolor{winbg}\textbf{26.7} \\
\quad AMC & 44.6 & 6.6 & 49.4 & 57.5 & \cellcolor{winbg}\textbf{62.0} \\
\quad MATH-500 & 69.8 & 15.6 & 79.9 & 84.8 & \cellcolor{winbg}\textbf{85.0} \\
\quad Minerva Math & 25.9 & 9.2 & 33.1 & \textbf{36.3} & 33.5 \\
\quad OlympiadBench & 35.8 & 4.6 & 39.4 & 48.9 & \cellcolor{winbg}\textbf{50.4} \\
\midrule
\textbf{Knowledge} & 48.5 & 4.3 & 55.4 & 56.9 & \cellcolor{winbg}\textbf{58.4} \\
\quad MMLU-Pro & 56.8 & 3.3 & \textbf{68.7} & 68.0 & \cellcolor{winbg}68.7 \\
\quad GPQA-Diamond & 40.2 & 5.3 & 42.2 & 45.7 & \cellcolor{winbg}\textbf{48.1} \\
\midrule
\textbf{Logical reasoning} & 78.1 & 25.0 & 94.7 & 90.8 & \cellcolor{winbg}\textbf{95.4} \\
\quad BBH Logical Deduction (3 obj.) & 83.6 & 35.2 & \textbf{98.6} & 96.4 & \cellcolor{winbg}97.8 \\
\quad BBH Logical Deduction (5 obj.) & 63.0 & 12.3 & \textbf{93.4} & 88.2 & \cellcolor{winbg}93.3 \\
\quad BBH Web of Lies & 87.7 & 27.6 & 92.2 & 87.9 & \cellcolor{winbg}\textbf{95.0} \\
\midrule
\textbf{Spatial \& tracking} & 59.0 & 17.6 & \textbf{85.6} & 83.4 & \cellcolor{winbg}84.2 \\
\quad BBH Tracking Shuffled Objects & 51.1 & 1.8 & \textbf{99.1} & 95.8 & 89.5 \\
\quad BBH Geometric Shapes & 38.0 & 3.4 & 60.1 & 58.2 & \cellcolor{winbg}\textbf{66.0} \\
\quad BBH Navigate & 87.9 & 47.7 & \textbf{97.7} & 96.2 & \cellcolor{winbg}97.2 \\
\midrule
\textbf{Average (13 benchmarks)} & 53.5 & 13.3 & 67.2 & 68.4 & \cellcolor{winbg}\textbf{70.3} \\
\bottomrule
\end{tabular}
\label{tab:full-qwen38bbase}
\end{table}

\begin{table}[htbp]
\centering\footnotesize
\setlength{\tabcolsep}{4pt}
\caption{Per-benchmark results for Mistral-Nemo-Instruct-2407 (pass@1 \%, $8{,}192$-token budget). Layout, shading, and bold as in Table~\ref{tab:general-reasoning}.}
\begin{tabular}{@{}l ccccc@{}}
\toprule
 &  & \multicolumn{3}{c}{Controls} & \gamesmith{} \\
\cmidrule(lr){3-5}\cmidrule(lr){6-6}
Benchmark & Base & Format & Non-game & SPIRAL & matched \\
\midrule
\textbf{Mathematics} & 16.1 & 2.1 & \textbf{19.6} & 12.3 & \cellcolor{winbg}17.2 \\
\quad AIME & 0.8 & 0.0 & 0.8 & 0.8 & \cellcolor{winbg}\textbf{3.3} \\
\quad AMC & \textbf{16.3} & 3.6 & 15.7 & 7.5 & \cellcolor{winbg}15.1 \\
\quad MATH-500 & 40.0 & 2.1 & \textbf{50.1} & 31.6 & \cellcolor{winbg}38.8 \\
\quad Minerva Math & 13.1 & 3.9 & \textbf{15.2} & 11.4 & \cellcolor{winbg}14.3 \\
\quad OlympiadBench & 10.6 & 1.2 & \textbf{16.5} & 10.3 & \cellcolor{winbg}14.4 \\
\midrule
\textbf{Knowledge} & 34.1 & 2.3 & \textbf{41.2} & 28.4 & \cellcolor{winbg}33.4 \\
\quad MMLU-Pro & 35.9 & 1.3 & \textbf{52.8} & 40.0 & \cellcolor{winbg}46.7 \\
\quad GPQA-Diamond & \textbf{32.2} & 3.3 & 29.7 & 16.9 & \cellcolor{winbg}20.1 \\
\midrule
\textbf{Logical reasoning} & 70.0 & 14.5 & \textbf{93.9} & 76.5 & \cellcolor{winbg}80.3 \\
\quad BBH Logical Deduction (3 obj.) & 85.2 & 30.0 & \textbf{98.9} & 91.7 & \cellcolor{winbg}96.0 \\
\quad BBH Logical Deduction (5 obj.) & 59.1 & 13.6 & \textbf{92.8} & 74.2 & \cellcolor{winbg}82.7 \\
\quad BBH Web of Lies & 65.8 & 0.0 & \textbf{90.1} & 63.7 & 62.1 \\
\midrule
\textbf{Spatial \& tracking} & 66.9 & 16.9 & \textbf{78.4} & 69.0 & \cellcolor{winbg}76.2 \\
\quad BBH Tracking Shuffled Objects & 83.1 & 22.3 & \textbf{99.8} & 91.6 & \cellcolor{winbg}95.1 \\
\quad BBH Geometric Shapes & 42.7 & 7.6 & 41.8 & 30.7 & \cellcolor{winbg}\textbf{46.8} \\
\quad BBH Navigate & 75.0 & 20.9 & \textbf{93.6} & 84.6 & \cellcolor{winbg}86.7 \\
\midrule
\textbf{Average (13 benchmarks)} & 43.1 & 8.4 & \textbf{53.7} & 42.7 & \cellcolor{winbg}47.9 \\
\bottomrule
\end{tabular}
\label{tab:full-mistralnemoinstruct2407}
\end{table}

\begin{table}[htbp]
\centering\footnotesize
\setlength{\tabcolsep}{4pt}
\caption{Per-benchmark results for GLM-4-9B-0414 (pass@1 \%, $8{,}192$-token budget). Layout, shading, and bold as in Table~\ref{tab:general-reasoning}.}
\begin{tabular}{@{}l ccccc@{}}
\toprule
 &  & \multicolumn{3}{c}{Controls} & \gamesmith{} \\
\cmidrule(lr){3-5}\cmidrule(lr){6-6}
Benchmark & Base & Format & Non-game & SPIRAL & matched \\
\midrule
\textbf{Mathematics} & 32.2 & 2.7 & 34.5 & 38.5 & \cellcolor{winbg}\textbf{40.3} \\
\quad AIME & 11.7 & 0.0 & 10.0 & 11.7 & \cellcolor{winbg}\textbf{15.8} \\
\quad AMC & 30.1 & 1.5 & 34.6 & 41.0 & \cellcolor{winbg}\textbf{42.5} \\
\quad MATH-500 & 64.8 & 5.9 & 71.0 & 73.0 & \cellcolor{winbg}\textbf{75.3} \\
\quad Minerva Math & 25.5 & 4.6 & 26.7 & \textbf{28.8} & 27.9 \\
\quad OlympiadBench & 29.1 & 1.6 & 30.1 & 38.3 & \cellcolor{winbg}\textbf{39.8} \\
\midrule
\textbf{Knowledge} & 48.1 & 12.4 & 48.1 & 48.2 & \cellcolor{winbg}\textbf{50.1} \\
\quad MMLU-Pro & 60.9 & 13.8 & \textbf{63.2} & 61.2 & \cellcolor{winbg}\textbf{63.2} \\
\quad GPQA-Diamond & 35.2 & 11.0 & 33.0 & 35.2 & \cellcolor{winbg}\textbf{37.0} \\
\midrule
\textbf{Logical reasoning} & 86.3 & 33.5 & 94.5 & 92.8 & \cellcolor{winbg}\textbf{95.0} \\
\quad BBH Logical Deduction (3 obj.) & 92.0 & 63.8 & \textbf{99.3} & 98.2 & \cellcolor{winbg}98.3 \\
\quad BBH Logical Deduction (5 obj.) & 68.0 & 35.5 & 88.5 & 87.1 & \cellcolor{winbg}\textbf{91.9} \\
\quad BBH Web of Lies & \textbf{98.9} & 1.1 & 95.6 & 93.0 & \cellcolor{winbg}94.9 \\
\midrule
\textbf{Spatial \& tracking} & 63.4 & 30.5 & 83.8 & 82.6 & \cellcolor{winbg}\textbf{86.4} \\
\quad BBH Tracking Shuffled Objects & 47.0 & 24.2 & \textbf{99.4} & 98.0 & 97.3 \\
\quad BBH Geometric Shapes & 53.9 & 9.6 & 55.7 & 57.7 & \cellcolor{winbg}\textbf{67.9} \\
\quad BBH Navigate & 89.3 & 57.8 & \textbf{96.4} & 92.2 & \cellcolor{winbg}93.9 \\
\midrule
\textbf{Average (13 benchmarks)} & 54.3 & 17.7 & 61.8 & 62.7 & \cellcolor{winbg}\textbf{65.1} \\
\bottomrule
\end{tabular}
\label{tab:full-glm49b0414}
\end{table}


\subsection{Diagnostics of the Format control}
\label{app:format}
The Format control of Table~\ref{tab:general-reasoning} scores below the base model on every model, by $16.7$ points on Qwen3-4B-Base and by $34.6$ to $40.2$ points on the other three. 
This control is derived mechanically from the \gamesmith{} distillation data by deleting the reasoning block of every response and keeping only the final \texttt{\textbackslash boxed} action, so that the two settings share the same $8{,}208$ records, the same eight epochs, and the same optimiser settings. 
Table~\ref{tab:format-diag} reports generation-level diagnostics for this control under the evaluation protocol of \S\ref{sec:setup} ($k=4$, $8{,}192$ maximum new tokens, temperature $0.6$, top-$p$ $0.95$). 
We measure two quantities, each macro-averaged over the $13$ benchmarks: the share of samples that open a \texttt{\textbackslash boxed\{\}} within the first $32$ characters (Answer-first), and the share of answered samples whose answer is wrapped in square brackets in the manner of a game action (Action-shaped). 
The Qwen3-4B-Base rows come from a re-run of its three settings under the same protocol, since the original runs kept no per-sample outputs. This re-run reproduces the means of Table~\ref{tab:general-reasoning} to within $0.4$ points, and the Avg.\ column keeps the values of Table~\ref{tab:general-reasoning}.

We observe that the Format control emits its answer as its first tokens. 
On the three larger models the answer frequently takes the shape of a game action, whereas on Qwen3-4B-Base the answer is a bare letter or number. 
These results indicate that supervising the format without the reasoning removes the base model's own reasoning rather than adding to it. 
The Format column therefore records the direction and mechanism of this loss rather than a level of ability, since several of the multiple-choice benchmarks fall to or below chance.

\begin{table}[t]
\centering\small
\caption{Generation-level diagnostics for the Format control, macro-averaged over the $13$ benchmarks of Table~\ref{tab:general-reasoning} (\%). Avg.\ is the 13-benchmark mean of Table~\ref{tab:general-reasoning}, and the remaining columns are defined in the text.}
\begin{tabular}{@{}ll ccc@{}}
\toprule
Model & Setting & Avg.\ (13) & Answer-first & Action-shaped \\
\midrule
Qwen3-4B-Base & Base    & 45.5 &  1.0 & 0.0 \\
 & Format  & 28.8 & 99.9 & 0.0 \\
 & \gamesmith{} & 62.5 &  0.0 & 0.0 \\
\midrule
Qwen3-8B-Base & Base    & 53.5 &  0.5 & 0.0 \\
 & Format  & 13.3 & 99.9 & 68.5 \\
 & \gamesmith{} & 70.3 &  0.0 & 0.0 \\
\midrule
Mistral-Nemo-Instruct-2407 & Base    & 43.1 &  0.0 & 0.0 \\
 & Format  &  8.4 & 99.6 & 68.3 \\
 & \gamesmith{} & 47.9 &  0.0 & 0.0 \\
\midrule
GLM-4-9B-0414 & Base    & 54.3 &  0.0 & 0.0 \\
 & Format  & 17.7 & 98.7 & 79.4 \\
 & \gamesmith{} & 65.1 &  0.0 & 0.0 \\
\bottomrule
\end{tabular}
\label{tab:format-diag}
\end{table}

\subsection{Estimator comparison with one episode per variant}
\label{app:group-size}
The self-play runs of Table~\ref{tab:pipelines} play every sampled variant $G=5$ times per optimiser step, so that both estimators see the same batches. In this section, we examine what remains of the difference between the two estimators when each variant is played once ($G=1$). With one episode per variant there is no group to average over: RAE reduces to its running mean per variant and seat, and \ourrl{} reduces to the prior $m_g$ of \eqref{eq:ourrl}, because $K_g'=0$ in \eqref{eq:ourrl} and the leave-one-out term vanishes. The comparison therefore isolates the two sources of \eqref{eq:ourrl} that do not depend on the current batch: a per-variant running mean against a prior fitted across the corpus.

Both runs start from the \gamesmith{} distillation checkpoint, use the same corpus, sampler, token budget, and learning rate as the main runs (Appendix~\ref{app:hparams}), and differ from them only in $G$. The batch of $128$ transitions is kept, so a $G=1$ step covers five times as many distinct variants as a $G=5$ step. Checkpoints are selected as in Table~\ref{tab:pipelines}, Table~\ref{tab:group-size} reports the results, and Figure~\ref{fig:rl-curves-g1} follows the Qwen3-4B-Base runs across the budget.

We make three observations. First, the prior alone retains the advantage of \ourrl{}: with $G=1$, \ourrl{} exceeds RAE on every model, by $1.6$ points on Qwen3-4B-Base, $1.0$ on Qwen3-8B-Base, $3.1$ on Mistral-Nemo-Instruct-2407, and $0.6$ on GLM-4-9B-0414, and it improves on the distilled model on every model, as with $G=5$. Second, \ourrl{} is insensitive to the group size, whereas RAE is not. Its $G=1$ scores are within $0.3$ points of its $G=5$ scores in Table~\ref{tab:pipelines} on every model, while RAE loses $1.8$ points on Mistral-Nemo-Instruct-2407 when each variant is played once. Third, Figure~\ref{fig:rl-curves-g1} shows on Qwen3-4B-Base the same behaviour over the budget as with $G=5$: \ourrl{} is above its starting point from the first evaluated checkpoint and ends $2.5$ points above it, whereas RAE peaks at step $20$, falls below the starting point for most of the budget, and recovers only towards the end. Since a $G=1$ step covers five times as many distinct variants, these results indicate that the prior, which is defined on every variant, is what allows the estimator to make use of the breadth of the corpus.
\begin{table}[h]
\centering\small
\setlength{\tabcolsep}{5pt}
\caption{Self-play with one episode per variant ($G=1$) on the \gamesmith{} corpus (13-benchmark mean pass@1 \%). Both runs start from the \gamesmith{} distillation checkpoint of Table~\ref{tab:pipelines} and differ only in the estimator. Bold marks the higher of the two RL runs in each row.}
\label{tab:group-size}
\begin{tabular}{@{}lccc@{}}
\toprule
Model & Distill. & +RL (RAE, $G=1$) & +RL (\ourrl{}, $G=1$) \\
\midrule
Qwen3-4B-Base & 62.5 & 63.4 & \textbf{65.0} \\
Qwen3-8B-Base & 70.3 & 71.8 & \textbf{72.8} \\
Mistral-Nemo-Instruct-2407 & 47.9 & 45.5 & \textbf{48.6} \\
GLM-4-9B-0414 & 65.1 & 64.7 & \textbf{65.3} \\
\bottomrule
\end{tabular}
\end{table}

\subsection{Evaluation curves during self-play RL}
\label{app:rl-curves}
\begin{figure}[t]
\centering
\begin{subfigure}[t]{\textwidth}
\centering
\includegraphics[width=\textwidth]{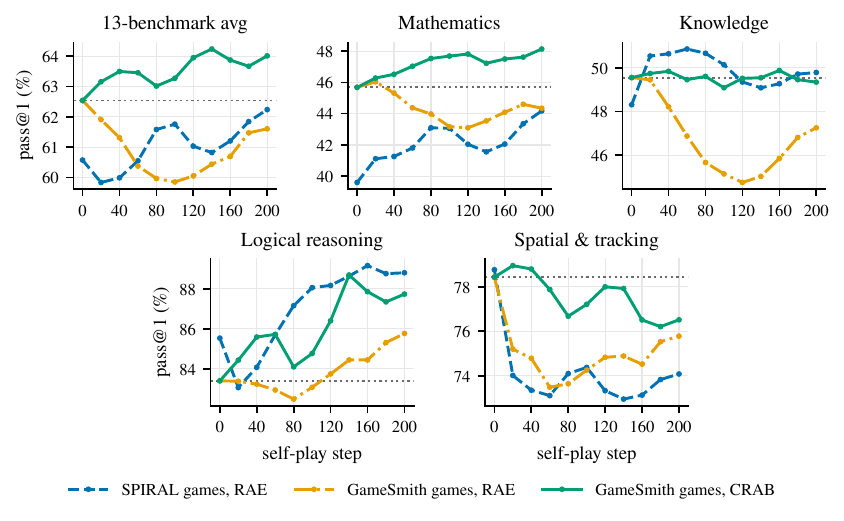}
\caption{Every sampled variant played $G=5$ times per step, the runs of Table~\ref{tab:pipelines}.}
\label{fig:rl-curves-g5}
\end{subfigure}
\par\medskip
\begin{subfigure}[t]{\textwidth}
\centering
\includegraphics[width=\textwidth]{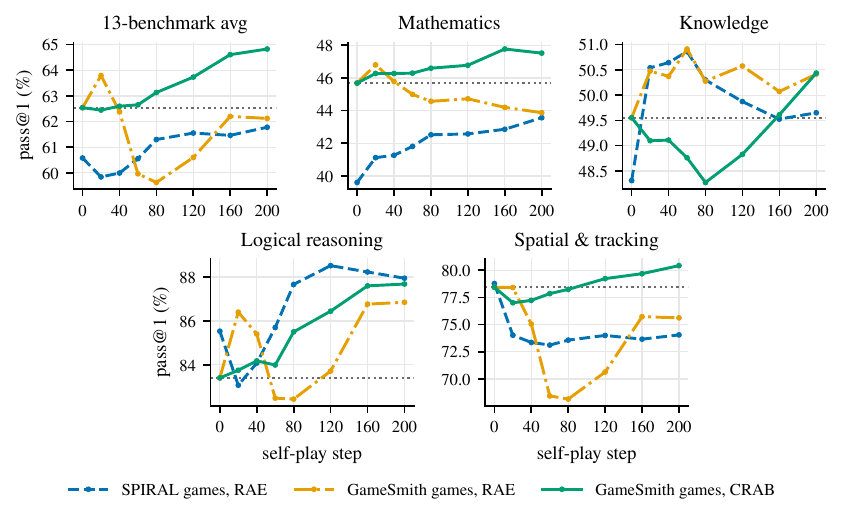}
\caption{Every sampled variant played once ($G=1$): the two \gamesmith{} runs of Appendix~\ref{app:group-size} and the SPIRAL run of Table~\ref{tab:pipelines}.}
\label{fig:rl-curves-g1}
\end{subfigure}
\caption{Self-play on Qwen3-4B-Base over the $200$-step budget, evaluated at the seven checkpoints of the grid of Table~\ref{tab:pipelines}, for the 13-benchmark mean and for each capability group (pass@1 \%). The dotted line is the \gamesmith{} distillation checkpoint from which the two \gamesmith{} pipelines start, and the SPIRAL pipeline starts from its own distillation checkpoint.}
\label{fig:rl-curves}
\end{figure}

Figure~\ref{fig:rl-curves} shows the self-play curves on Qwen3-4B-Base, and its top panel (Figure~\ref{fig:rl-curves-g5}) follows the three self-play pipelines of Table~\ref{tab:pipelines} across the whole budget. We make three observations. First, the \ourrl{} pipeline stays above its distillation checkpoint at every evaluated step, and its gain is concentrated on mathematics and logical reasoning, where it ends $2.3$ and $4.6$ points above the starting point. Second, RAE on the same corpus and checkpoint falls below the starting point by step $40$ and returns to it only at the end of the budget. The fall is driven by the knowledge group, where it loses more than five points by step $120$ before recovering. This is the behaviour anticipated in \S\ref{sec:intro} and \S\ref{sec:method:ourrl}: a baseline kept separately for each variant is unreliable when a run visits each variant only a few times, so self-play with RAE on the synthesised corpus is harmful for most of the budget, and the two RAE pipelines end level in Table~\ref{tab:pipelines} although the \gamesmith{} one starts $1.9$ points higher, whereas the SPIRAL pipeline, which replays three games, gains steadily from its lower start. Third, the two groups that the games exercise least are the least affected by the estimator: knowledge moves by at most two and a half points under the \ourrl{} and SPIRAL pipelines, and spatial and state tracking fluctuates around the starting point under \ourrl{} and falls below it under RAE on either corpus. The picture is similar when every sampled variant is played once: Figure~\ref{fig:rl-curves-g1} repeats the comparison at $G=1$ with the runs of Appendix~\ref{app:group-size} and shows the same ordering of the three pipelines.

\subsection{Draws and the reward signal}
\label{app:draws}
A zero-sum game can end in a draw, and a draw returns $0$ to both seats. A natural concern is that self-play drifts into draws and thereby loses its reward signal, as happens on Tic-Tac-Toe once both sides play well enough. Two mechanisms limit this effect, and Table~\ref{tab:draws} reports what happened in the runs of Table~\ref{tab:pipelines}.

The first mechanism is admission. The fingerprint of random play excludes a game that produces only draws, and the playability test requires at least $25\%$ of episodes between two score-maximising reference bots to be decisive (Appendix~\ref{app:verify}). In the admitted corpus, the base game of a family is decisive in at least $69\%$ of random episodes (\texttt{crossl}), in at least $90\%$ for $26$ of the \gsNfamilies{} families, and the median is $1.00$. Among the \gsNvariants{} admitted variants, three rule variants draw in nearly every random episode because the pattern that decides the game is rarely completed by random play, and one of them, the \texttt{nosqre} variant with a $3\times3$ losing square, is included in the self-play pool.

The second mechanism is the estimator. A drawn episode is not a lost sample: its return is $0$, and \eqref{eq:pg} uses $R(\tau_i)-b_i$, which is nonzero whenever the baseline is not exactly $0$. Since \ourrl{} shrinks the leave-one-out mean of the same game towards the prior $m_g$, $b_i$ differs from $R(\tau_i)$ unless the prior happens to equal it. Table~\ref{tab:draws} reports the fraction of episodes that were drawn, averaged over the first and the last $20$ steps of the budget.

Three observations stand out. 
First, on the synthesised corpus the draw rate stays between $3\%$ and $15\%$ throughout training on every model, and on Qwen3-4B-Base it decreases under \ourrl{}. Second, on the three SPIRAL games it climbs from about $10\%$ to $25$ to $31\%$ by the end of the budget on every model, which is the pattern expected of a small fixed set of games that the policy learns to play to a draw. The concern is therefore real for hand-written games, and it is precisely what the breadth of the corpus removes. 
Third, we also logged, for the \ourrl{} runs, whose baseline uses the group of episodes of the same variant, the fraction of groups of size at least two whose returns were all identical: it lies between $4$ and $13\%$ across the four models. Even in these groups no advantage is exactly zero (the logged fraction of advantages equal to zero is $0.0$ in every run), because the prior term keeps the baseline away from the group's common return, and the state-value term of \eqref{eq:ourrl-turn} adds a per-decision signal on top. 
A nonzero advantage is a necessary condition for a gradient signal, not a sufficient one. 
If every action available in a group leads to the same return, the expected gradient of that group is zero for any action-independent baseline, and a nonzero baseline only adds a zero-mean term to the update. The quantities that bear on the signal are therefore the draw rate and the fraction of same-return groups, which bound the share of the batch that can carry no signal, whereas the absence of zero advantages rules out only the degenerate case of a batch of draws that contributes nothing at all to the gradient. Whether the remaining advantages carry signal is a question about policy improvement, not about the advantages themselves.

\begin{table}[h]
\centering\footnotesize
\setlength{\tabcolsep}{3pt}
\caption{Draw rate during self-play, from the training logs of the runs of Table~\ref{tab:pipelines}, steps $1$ to $200$.}
\begin{tabular}{@{}lllrr@{}}
\toprule
Model & Games & Estimator & \multicolumn{2}{c}{Draw rate (\%)} \\
\cmidrule(lr){4-5}
 & & & first $20$ steps & last $20$ steps \\
\midrule
Qwen3-4B-Base & SPIRAL & RAE & 7.1 & 25.0 \\
 & \gamesmith{} & RAE & 5.7 & 10.9 \\
 & \gamesmith{} & \ourrl{} & 7.4 & 5.3 \\
\midrule
Qwen3-8B-Base & SPIRAL & RAE & 9.5 & 30.7 \\
 & \gamesmith{} & RAE & 4.2 & 15.2 \\
 & \gamesmith{} & \ourrl{} & 4.3 & 14.2 \\
\midrule
Mistral-Nemo-Instruct-2407 & SPIRAL & RAE & 12.7 & 29.4 \\
 & \gamesmith{} & RAE & 3.4 & 10.8 \\
 & \gamesmith{} & \ourrl{} & 3.6 & 11.8 \\
\midrule
GLM-4-9B-0414 & SPIRAL & RAE & 9.9 & 30.7 \\
 & \gamesmith{} & RAE & 3.0 & 12.5 \\
 & \gamesmith{} & \ourrl{} & 2.7 & 11.1 \\
\bottomrule
\end{tabular}
\label{tab:draws}
\end{table}

\subsection{Win rates on held-out variants}
\label{app:game-strength}
The benchmarks of Table~\ref{tab:pipelines} measure transfer. In this section, we examine whether the self-play runs also play the games better. We compare the two pipelines that differ in both the corpus and the estimator, namely SPIRAL games with RAE and \gamesmith{} games with \ourrl{}, at the checkpoints selected for Table~\ref{tab:pipelines} (Table~\ref{tab:game-strength}).

The test set consists of $54$ variants that neither pipeline was trained on: for $18$ of the \gsNfamilies{} families, one variant on each of the encoding, rule, and size axes of Appendix~\ref{app:corpus}. 
Every variant is played for $40$ episodes, $20$ from each seat, with the same budget and sampling as training. 
The opponent is Qwen3-4B-Instruct under the same budget, and the two pipelines are also played against each other on the same positions. 
An invalid move or a reply that runs to the budget counts as a loss. The $40$ episodes of a variant form $13$ best-of-three series, and each cell reports the share of series won, so that $50$ corresponds to parity with the opponent.

Table~\ref{tab:game-strength-rl} shows that the \gamesmith{} pipeline plays better than the SPIRAL pipeline. Against Qwen3-4B-Instruct it wins $42.5\%$ of the series overall against $33.5\%$, and head-to-head it wins $62.7\%$. The gap is concentrated on the grid genre, where the \gamesmith{} pipeline wins $40.6\%$ of the series against $18.4\%$: the SPIRAL pipeline loses a quarter of its grid games by moving to an occupied cell or by running past the budget, both of which count as losses, and it wins only $26\%$ of the head-to-head series on grid. The two pipelines are level on negotiation, and the gap on card is within a few points. Table~\ref{tab:game-strength-sft} repeats the protocol on the two distillation checkpoints from which the pipelines start, and the ordering is the same before any self-play: distillation from the synthesised games plays better than distillation from the three SPIRAL games on every genre, by the largest margin on grid. The ordering of the self-play pipelines is therefore inherited from their starting points and preserved by the self-play stage.

\begin{table}[h]
\centering\small
\setlength{\tabcolsep}{5pt}
\caption{Playing strength on the held-out variants, as the share of best-of-three series won (\%). The first four columns play against Qwen3-4B-Instruct under the training budget, and the last column of (b) plays the two self-play pipelines against each other. An invalid move or a reply that runs to the token budget counts as a loss. Bold marks the higher value in each column.}
\label{tab:game-strength}
\begin{subtable}{\textwidth}
\centering
\caption{The two distillation checkpoints from which the self-play pipelines start, before any self-play.}
\label{tab:game-strength-sft}
\begin{tabular}{@{}lcccc@{}}
\toprule
 & \multicolumn{4}{c}{vs. Qwen3-4B-Instruct} \\
Distilled on & Card & Grid & Negotiation & All \\
\midrule
SPIRAL games & 34.1 & 12.8 & 32.7 & 26.7 \\
\gamesmith{} games & \textbf{40.1} & \textbf{36.8} & \textbf{43.6} & \textbf{39.7} \\
\bottomrule
\end{tabular}
\vspace{0.5cm}
\end{subtable}
\begin{subtable}{\textwidth}
\centering
\caption{The self-play pipelines of Table~\ref{tab:pipelines} at their selected checkpoints.}
\label{tab:game-strength-rl}
\begin{tabular}{@{}lccccc@{}}
\toprule
 & \multicolumn{4}{c}{vs. Qwen3-4B-Instruct} & Head-to-head \\
Self-play pipeline & Card & Grid & Negotiation & All & series won \\
\midrule
SPIRAL games, RAE (step 200) & 36.2 & 18.4 & \textbf{51.0} & 33.5 & 37.3 \\
\gamesmith{} games, \ourrl{} (step 120) & \textbf{40.7} & \textbf{40.6} & 48.7 & \textbf{42.5} & \textbf{62.7} \\
\midrule
Series per column & 312 & 234 & 156 & 702 & 702 \\
\bottomrule
\end{tabular}
\end{subtable}
\end{table}

\FloatBarrier  
\section{Synthesis prompts}
\label{app:prompts}
The designer and the engineer are the same generator model, namely GPT-5 accessed through its API with the provider's default sampling. The designer receives up to $32{,}768$ completion tokens and the engineer up to $24{,}576$. An engineer implementation that fails a replay test receives the test output verbatim and is revised, for at most five attempts. The system prompts of both roles for the three templates, the user messages that instantiate them, and the playtest prompt are reproduced below verbatim from the synthesis code. Braces mark values that are filled in at run time.

\begin{promptbox}{Designer system prompt, grid-placement template}
You are a game designer for an RL research pipeline. Propose NEW two-player, zero-sum, turn-based GRID 
  PLACEMENT games implementable in a fixed skeleton (players alternate placing marks on a grid; game 
  logic only controls: which cells are legal, what placing does, and when someone wins/loses/draws). The
   games must be DIFFERENT from standard Tic-Tac-Toe and from each other, deterministic, 
  perfect-information, and end within ~50 moves.

Good mechanic directions: misere/anti variants, Notakto, Order&Chaos, Domineering-style blocking, 
  connection goals, custodial capture, majority-of-lines scoring, forbidden patterns.

HARD RULE: define player roles and objectives by MOVE ORDER (the first player / the second
player), NEVER by display symbols or labels -- those are cosmetic and get randomized.

Return a JSON list; each item:
{"key": "SHORTKEY (<=6 uppercase letters)",
 "name": "...",
 "rules": "complete, unambiguous natural-language rules a strong player could play from",
 "semantic_knobs": {"knob_name": "meaning + sensible int/bool values"},
 "deduction_note": "why strong play requires verifiable step-by-step deduction (1-2 sentences)"}
\end{promptbox}

\begin{promptbox}{Designer system prompt, betting-with-showdown template}
You are a game designer for an RL research pipeline. Propose NEW two-player, zero-sum BETTING/SHOWDOWN 
  card games implementable in a fixed skeleton. The skeleton is sealed: it deals each player hidden 
  cards from a deck of ranks, runs a standard limit-betting round (check/bet/call/fold/raise), and 
  settles chips. THE ONLY THING EACH GAME CONTROLS IS THE HAND RANKING FUNCTION: given a player's cards,
   compute a comparable strength value (higher wins the showdown; ties split). Games must differ from 
  plain Kuhn poker (highest card wins) and from each other via genuinely different ranking logic.

Good directions: lowball, middle-card-wins, sum modulo N, pair/duplicate bonuses, distance between 
  cards, parity-based rankings, "closest to target sum without exceeding it".
Knobs may include deck size, hand size, ante, bet size, plus ranking-specific parameters.

HARD RULE: define player roles and objectives by MOVE ORDER (the first player / the second
player), NEVER by display symbols or labels -- those are cosmetic and get randomized.

Return a JSON list; each item:
{"key": "SHORTKEY (<=6 uppercase letters)",
 "name": "...",
 "rules": "complete, unambiguous natural-language rules a strong player could play from",
 "semantic_knobs": {"knob_name": "meaning + sensible int/bool values"},
 "deduction_note": "why strong play requires verifiable step-by-step deduction (1-2 sentences)"}
\end{promptbox}

\begin{promptbox}{Designer system prompt, resource-negotiation template}
You are a game designer for an RL research pipeline. Propose NEW two-player RESOURCE-NEGOTIATION games 
  implementable in a fixed skeleton that generalizes SimpleNegotiation. The skeleton is sealed: it deals
   each player quantities of several resource types plus a PRIVATE per-unit value for each type; players
   alternate turns; on your turn you either propose ONE trade from a listed menu of concrete swaps, or 
  accept/refuse the opponent's pending proposal; accepted trades move the goods. Settlement is ZERO-SUM:
   whoever's inventory SCORE grew more by the final turn wins (equal growth = draw).

THE ONLY THING EACH GAME CONTROLS IS HOW AN INVENTORY BECOMES A SCORE (and optionally which swaps are on
   the menu). Plain SimpleNegotiation scores an inventory as sum(quantity x private unit value); your 
  games must differ via genuinely different scoring structure, so that which trades are worth making 
  changes qualitatively.

Good directions: diminishing returns (each extra unit of a type is worth less), SET-COLLECTION bonuses 
  (points for holding at least k of every type), complementary pairs (a bonus per matched pair across 
  two specific types), threshold/quota bonuses (a jump once a type crosses a level), monopoly bonuses 
  (holding the majority of a type), caps (units beyond a cap score nothing), parity or spread bonuses 
  (rewarding balance across types instead of hoarding one).

CRITICAL: the game must stay non-degenerate under RATIONAL play -- mutually beneficial trades must EXIST
   (both sides can gain because their private values and holdings differ), otherwise a rational pair 
  never trades and every game is a draw. Say in the rules why a rational opponent would ever accept a 
  proposal.
Knobs may include resource count, turn budget, valuation spread, trade scale, plus scoring-specific 
  parameters (bonus size, threshold level, set size, cap).

HARD RULE: define player roles and objectives by MOVE ORDER (the first player / the second
player), NEVER by display symbols or labels -- those are cosmetic and get randomized.

Return a JSON list; each item:
{"key": "SHORTKEY (<=6 uppercase letters)",
 "name": "...",
 "rules": "complete, unambiguous natural-language rules a strong player could play from",
 "semantic_knobs": {"knob_name": "meaning + sensible int/bool values"},
 "deduction_note": "why strong play requires verifiable step-by-step deduction (1-2 sentences)"}
\end{promptbox}

\begin{promptbox}{Designer user message}
Propose {n} games. Existing corpus already has: {list of admitted games on this template}. Avoid 
  near-duplicates of these. Output ONLY the JSON array (in a ```json fence), no prose.
\end{promptbox}

\begin{promptbox}{Engineer system prompt, grid-placement template}
You implement a two-player grid game inside a FIXED skeleton. You may ONLY write:
1) a frozen dataclass `Config(GridConfig)` adding the game's semantic knobs (with defaults);
2) a class `Game(GridGameBase)` implementing ONLY these hooks (all in CANONICAL coordinates):
   - legal_moves(self, board, pid, extras) -> list[(r,c)]   (optional; default = empty cells)
   - apply_move(self, board, pid, r, c, extras) -> None      (optional; default = board[r][c]=pid)
   - outcome(self, board, pid, r, c, extras) -> None | ("win", pid) | ("lose", pid) | ("draw",)
     called right after pid placed at (r,c); ("lose", pid) means the MOVER loses.
   - rules_text(self, cfg, player_id, symbols) -> str        (clear rules; mention both symbols)
   - init_extras(self) -> dict                               (optional per-episode state)
3) a `make_suite(cfg_cls)` returning {"train": [cfgs], "val": [cfgs], "test": [cfgs]} where train
   uses randomize_codec=True and test holds out symmetry="transpose" plus one rule-knob variant;
4) the closing `globals().update(family_boilerplate(KEY, __name__, Game, Config,
   semantic_fields=(...), suite_fn=make_suite))`.
Board stores None / 0 / 1 (player ids) ONLY -- never symbols. Do not touch rendering, parsing,
codecs, textarena, or the State machinery. The game must always terminate (the skeleton
auto-draws when neither player has legal moves).
HARD RULES: deterministic logic only (no randomness in hooks, no I/O, no clock). Imports allowed: 
  dataclasses, typing, math, itertools, and the skeleton import shown above. Hooks COMPUTE, they never 
  render or print. rules_text must describe the rules using the provided labels argument, never 
  hardcoded display strings. Your Config MUST override structural defaults (board/deck/hand/slot sizes) 
  to the values the SPEC intends -- do NOT inherit skeleton defaults; the game's mechanic must be 
  non-degenerate at the Config defaults (a behavioral-duplicate gate will reject games that collapse to 
  an existing ordering). Output ONE ```python block, nothing else.
\end{promptbox}

\begin{promptbox}{Engineer system prompt, betting-with-showdown template}
You implement a betting/showdown card game inside a FIXED skeleton. You may ONLY write:
1) a frozen dataclass `Config(CardConfig)` adding the game's semantic knobs (with defaults);
2) a class `Game(CardGameBase)` implementing ONLY these hooks:
   - hand_value(self, hand, cfg) -> comparable value (int or tuple; HIGHER wins the showdown,
     equal values split the pot). `hand` is a list of int ranks in [0, cfg.n_ranks).
   - rules_text(self, cfg, player_id, labels) -> str. `labels` is a dict with the current
     display words for actions ('check','bet','call','fold','raise') and labels['ranks'] =
     rank labels low-to-high. Explain the ranking rule precisely using these labels.
3) a `make_suite(cfg_cls)` returning {"train": [cfg_cls(randomize_codec=True)],
   "val": [cfg_cls(rank_scheme="number", action_scheme="formal")],
   "test": [cfg_cls(rank_scheme="letter", action_scheme="casual"), <one rule-knob variant>]};
4) the closing `globals().update(card_family_boilerplate(KEY, __name__, Game, Config,
   semantic_fields=(...), suite_fn=make_suite))`.
Import: from driftgym.skeletons.card_game import CardConfig, CardGameBase, card_family_boilerplate
Do not touch dealing, betting flow, chips, rendering, or State machinery.
HARD RULES: deterministic logic only (no randomness in hooks, no I/O, no clock). Imports allowed: 
  dataclasses, typing, math, itertools, and the skeleton import shown above. Hooks COMPUTE, they never 
  render or print. rules_text must describe the rules using the provided labels argument, never 
  hardcoded display strings. Your Config MUST override structural defaults (board/deck/hand/slot sizes) 
  to the values the SPEC intends -- do NOT inherit skeleton defaults; the game's mechanic must be 
  non-degenerate at the Config defaults (a behavioral-duplicate gate will reject games that collapse to 
  an existing ordering). Output ONE ```python block, nothing else.
\end{promptbox}

\begin{promptbox}{Engineer system prompt, resource-negotiation template}
You implement a two-player resource-NEGOTIATION game inside a FIXED skeleton. You may ONLY write:
1) a frozen dataclass `Config(NegConfig)` adding the game's semantic knobs (with defaults) AND
   overriding structural knobs (n_resources, max_turns, qty_min, qty_max) to what the SPEC intends;
2) a class `Game(NegGameBase)` implementing ONLY these hooks (all COMPUTE-only, never render):
   - inventory_score(self, quantities, values, cfg) -> int   THE MECHANIC. `quantities` is a list of
     ints indexed by canonical resource index (0=cheapest .. n-1=dearest base value) and `values` is
     that player's private per-unit values, same indexing. Return an int score. Default = sum(q*v).
     Must be deterministic and pure. Higher = better for that player.
   - menu(self, cfg) -> list[(give_idx, get_idx, give_qty, get_qty)]   (optional) the proposable
     swaps, in a STABLE order (canonical actions are `offer {k}` by list index). The skeleton default
     lists every ordered resource pair at BASE-VALUE PARITY (gcd ratios, e.g. 8 units of a 5-value
     resource for 1 unit of a 40-value one). KEEP PARITY-LIKE RATIOS: equal-quantity swaps across an
     unequal value ladder make every proposer-favourable trade harmful to the responder, so a
     rational pair never trades and the game degenerates to a guaranteed draw.
     WARNING: HARD GATE (certificate C-rational): a reference bot that greedily maximizes ITS OWN score
     plays both seats; the game FAILS unless it executes >=0.5 trades/episode and >=25% of episodes
     are decisive. The skeleton's DEFAULT menu is parity in the *base* values, so if your
     inventory_score is strongly non-linear (steep diminishing returns, big set/quota/variance
     bonuses) those swaps are LOSS-MAKING for both sides under your own scoring and nothing ever
     trades. Two ways to satisfy the gate: (i) keep the non-linearity MILD so that score stays close
     to monotone in base value (bonuses small relative to sum(q*v)), or (ii) override menu() so the
     quantities are parity under YOUR scoring rule at typical inventories. State in rules_text why a
     rational opponent would accept.
   - trade_ok(self, give_idx, get_idx, gq, rq, quantities, cfg) -> bool   (optional; default = the
     proposer holds at least gq of give_idx)
   - rules_text(self, cfg, player_id, labels) -> str   describe the scoring rule precisely using
     labels['resources'] (display names), labels['base_values'], and labels['offer'/'accept'/'deny'].
3) a `make_suite(cfg_cls)` returning {"train": [cfg_cls(randomize_codec=True), ...],
   "val": [cfg_cls(resource_scheme="num", action_scheme="alt")],
   "test": [cfg_cls(resource_scheme="minerals", action_scheme="market"), <one rule-knob variant>,
            <one structural variant>]};
4) the closing `globals().update(neg_family_boilerplate(KEY, __name__, Game, Config,
   semantic_fields=(...), suite_fn=make_suite))`.
Import: from driftgym.skeletons.negotiation_game import NegConfig, NegGameBase, neg_family_boilerplate
Settlement is zero-sum and owned by the skeleton (it compares each side's score growth) -- do not
touch dealing, the offer state machine, trade execution, rendering, or State machinery.
HARD RULES: deterministic logic only (no randomness in hooks, no I/O, no clock). Imports allowed: 
  dataclasses, typing, math, itertools, and the skeleton import shown above. Hooks COMPUTE, they never 
  render or print. rules_text must describe the rules using the provided labels argument, never 
  hardcoded display strings. Your Config MUST override structural defaults (board/deck/hand/slot sizes) 
  to the values the SPEC intends -- do NOT inherit skeleton defaults; the game's mechanic must be 
  non-degenerate at the Config defaults (a behavioral-duplicate gate will reject games that collapse to 
  an existing ordering). Output ONE ```python block, nothing else.
\end{promptbox}

\begin{promptbox}{Engineer user message (first attempt)}
SKELETON SOURCE (read the base class contract):
```python
{template source file}
```

REFERENCE IMPLEMENTATION on this skeleton:
```python
{hand-written seed game of the template}
```

Now implement THIS game. KEY={KEY}. module will be driftgym.corpus.{key}.env
SPEC:
{designer specification as JSON}
\end{promptbox}

\begin{promptbox}{Engineer user message (repair, appended after each failed attempt)}
Attempt {k} FAILED:
{verbatim output of the failing replay tests, or the traceback}

Fix the implementation. Output the complete corrected file as ONE ```python block.
\end{promptbox}

\begin{promptbox}{Independent re-implementation user message (same engineer system prompt; the first implementation is not shown)}
SKELETON SOURCE (read the base class contract):
```python
{template source file}
```

Now implement this game FROM ITS SPEC alone. KEY={KEY}T. module will be driftgym.corpus.{key}.twin_env
Use EXACTLY these Config defaults (field names and values):
{Config defaults of the first implementation as JSON}

SPEC:
{designer specification}
\end{promptbox}

\begin{promptbox}{Playtest system prompt (the model plays one seat; the user message is the rules text and the current observation)}
You are playing a two-player game. Read the rules and current state, then move.
Reply with EXACTLY one line: [label] | END: win  or  [label] | END: lose  or
[label] | END: draw  or  [label] | END: none
END states what happens to YOU immediately after this move by the rules: win/lose/draw if the game ends 
  now, none if it continues. Choose the strongest legal move.
\end{promptbox}

\end{document}